\documentclass{article}

\usepackage[preprint]{neurips_2026}

\usepackage[utf8]{inputenc} 
\usepackage[T1]{fontenc}    
\usepackage{hyperref}       
\usepackage{url}            
\usepackage{booktabs}       
\usepackage{amsfonts}       
\usepackage{nicefrac}       
\usepackage{microtype}      
\usepackage[table]{xcolor}  

\usepackage{amsmath}
\usepackage{algorithm}
\usepackage{algpseudocode}
\usepackage{multirow}
\usepackage{graphicx}

\title{Function-Aware Fill-in-the-Middle as Mid-Training for Coding Agent Foundation Models}

\author{
\textbf{Yubo Wang}$^{1,5*}$ \quad \textbf{Jiarong Liang}$^{1*}$ \quad \textbf{Yuxuan Zhang}$^{2}$ \quad \textbf{Xuye Liu}$^{1}$ \quad \textbf{Cong Wei}$^{1,3}$ \\
    \textbf{Yuyu Zhang}$^{4}$ \quad \textbf{Ping Nie}$^{1}$ \quad \textbf{Wenhu Chen}$^{1,5}$ \\[4pt]
        $^{1}$University of Waterloo, $^{2}$University of British Columbia, $^{3}$NVIDIA, $^{4}$Verdent AI, $^{5}$Vector Institute
      }

\begin{document}

\maketitle

{\renewcommand{\thefootnote}{*}\footnotetext{Equal contribution. Corresponding authors: yubo.wang.sunny@gmail.com, hustchenwenhu@gmail.com}}

\vspace{-3.5ex}
\begin{center}
\vspace{-3ex}
\small \url{https://github.com/TIGER-AI-Lab/FIM-Midtraining}
\end{center}

\begin{abstract}
Coding agents must integrate external tool returns into ongoing reasoning---a
capability that standard left-to-right pretraining on code exposes only in its
forward direction. We observe that the
\textit{action $\rightarrow$ observation $\rightarrow$ continuation} loop of a
coding agent is structurally isomorphic to a function call site, where a
caller binds arguments, a callee returns a value computed elsewhere, and
downstream code consumes that value. This conditioning structure exists at
internet scale in ordinary code. We exploit it through
\emph{function-aware fill-in-the-middle (FIM) mid-training}: a self-supervised
objective that masks functions selected via program dependency graph
analysis and a complexity--inferability double criterion. We mid-train
Qwen2.5-Coder-Instruct (7B/14B) and Qwen3-8B on a 2.6B-token decontaminated
corpus drawn from 968 GitHub repositories, then apply existing agentic
post-training pipelines.
Mid-training improves SWE-Bench-Verified by $+2.8/+3.0$ at 7B/14B and by
$+3.2$ on Qwen3-8B; SWE-Bench-Lite gains are $+3.7/+4.0/+5.4$ on the same
models. The improvement holds across two post-training pipelines (R2E-Gym,
SWE-Smith) and on a non-Qwen2.5 base (Qwen3-8B with SWE-Lego). Beyond
in-domain gains, mid-training also mitigates the capability erosion that
agentic post-training otherwise inflicts on non-agent coding (e.g.,
LiveCodeBench) and non-coding tool-use benchmarks ($\tau$-bench, BFCL):
although the mid-training corpus contains Python code only, the
function-call inductive bias survives post-training and yields consistent gains.
\end{abstract}

\vspace{-2ex}

\begin{figure}[h!]
\centering
\includegraphics[width=0.9\linewidth]{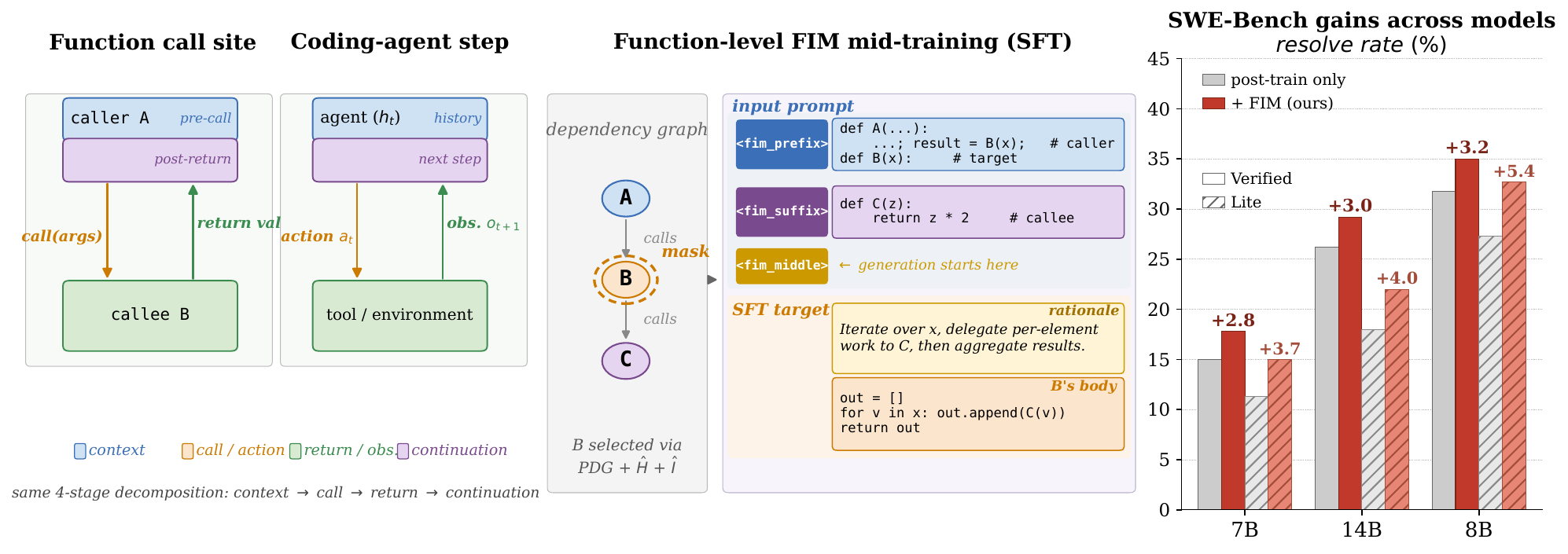}

\vspace{-1.9ex}
\caption{%
\textbf{Left:} A function call site and a single step of a coding agent
are \emph{structurally similar}, decomposing into the
same four stages: context, call/action, return/observation, continuation.
\textbf{Middle:} We exploit this analogy via function-aware FIM
mid-training. A function $B$ is selected from the program dependency
graph using complexity ($\hat{H}$) and inferability ($\hat{I}$) scores;
the model is then mid-trained to fill in $B$'s body together with a
CoT rationale, given the surrounding file as an
FIM-formatted prompt.
\textbf{Right:} Mid-training yields consistent gains across both Qwen2.5-Coder-Instruct (7B, 14B) and Qwen3 (8B) on SWE-Bench-Verified (solid bars) and SWE-Bench-Lite (hatched bars).}
\vspace{-1.5ex}
\label{fig:teaser}
\end{figure}

\section{Introduction}
\label{sec:intro}

Coding agents that resolve real software engineering issues have moved from
research demos to deployed systems~\citep{swebench, sweagent,
openhands}. Their progress, however, has been driven almost entirely by
scaling synthetic agent-trajectory data during post-training: pipelines such
as SWE-Gym~\citep{swegym}, R2E-Gym~\citep{r2egym}, and
SWE-Smith~\citep{swesmith} curate or synthesize trajectories that
imitate human or LLM behavior on issue-resolution tasks. The base model
these pipelines start from is typically a code LLM trained with next-token
prediction (and, in some cases, random-span FIM) on internet-scale
code~\citep{qwencoder, deepseekcoder, starcoder}. Between
these two stages lies a training-time gap: the base model is rarely
optimized for the conditioning structure that agentic post-training will
later demand. We treat this gap as an opportunity for a dedicated
\emph{mid-training} stage that aligns the base model with agent-relevant
inductive biases before agent-specific data is introduced.

Our central observation is that the inductive bias required by a coding
agent already exists in ordinary code, but in a shape that left-to-right
pretraining systematically under-exposes. At each step an agent maintains a
history $h_t$, samples an action $a_t \sim \pi(a_t \mid h_t)$, receives an
observation $o_{t+1}$ produced by an external process, and continues
conditioned on the entire trace. This four-part decomposition---context,
action, externally-computed return, continuation---is precisely the
decomposition of a function call site: pre-call code that establishes intent
and binds arguments; the call itself; a return value produced by code
outside the immediate scope; and downstream code that consumes the return
value (Figure~\ref{fig:teaser}, left). A model trained to reason
\emph{bidirectionally} about function-level dependencies must learn to
reconstruct a callee's behavior from caller context and downstream usage,
which is the same competence required to predict an agent's continuation
given a history and a tool return. The correspondence is structural rather
than literal---FIM training conditions on a given suffix while agent
inference generates one---but it suggests that representations induced by
the former should transfer to the latter, an empirical question we address
in Section~\ref{sec:exp}.

The fill-in-the-middle objective is not new~\citep{bavarian2022fim};
recent code LLMs mix random-span FIM into pretraining~\citep{qwencoder,
deepseekcoder, starcoder}. Random-span FIM is nonetheless poorly aligned
with agentic conditioning for three reasons. \emph{(i) Span boundaries
are syntactically arbitrary:} most masked spans cut through expressions
or partial statements and carry weak signal about function-level
dependencies. \emph{(ii) There is no reasoning supervision:} the model
fills the span directly, with no intermediate rationale mirroring an
agent's think-then-act pattern. \emph{(iii) The objective is dissolved
into pretraining:} by the time post-training begins, any FIM-conferred
structural prior has been amortized across trillions of unrelated tokens.
We address all three points: masking targets are selected at function
granularity via program dependency graph analysis with a
base-model-agnostic complexity--inferability double criterion
(Section~\ref{sec:method:selection}); chain-of-thought rationales are
embedded \emph{inside} the FIM middle span so the model produces
reasoning consistent with the eventual code
(Section~\ref{sec:method:cot}); and the objective is applied at a
dedicated mid-training stage, concentrating its signal immediately before
agentic post-training.

We evaluate this recipe along three axes of robustness. \emph{On the
Qwen2.5-Coder-Instruct series}, mid-training improves SWE-Bench-Verified
by $+2.8$ and $+3.0$ points at 7B and 14B respectively, indicating that
the structural prior is not absorbed by larger pretrained models in the
practical deployment range. \emph{Across two post-training pipelines},
mid-training improves both R2E-Gym and SWE-Smith on the same 7B base,
with the SWE-Smith pairing yielding $+5.3$ points on SWE-Bench-Verified.
\emph{On a non-Qwen2.5 base}, mid-training transfers to Qwen3-8B (paired
with SWE-Lego) for a $+3.2$-point gain on SWE-Bench-Verified; this single
comparison varies the post-training pipeline jointly with the base model
and should be read as evidence that the prior is not specific to a single
Qwen2.5-Coder $+$ R2E-Gym/SWE-Smith combination, rather than as a
guarantee across families.

A second set of results probes our motivating hypothesis. All checkpoints
are evaluated after the full pipeline (post-training alone, or
mid-training followed by post-training), since FIM-only checkpoints
degrade instruction-following and are not comparable to instruction-tuned
baselines. Agentic post-training alone substantially regresses non-target
capabilities at 14B, dropping LiveCodeBench, BFCL, and $\tau$-bench by
double-digit margins in some cases; adding mid-training before the same
post-training restores $+11.1$ on LiveCodeBench, $+2.4$ on BFCL, and
$+3.9$ on $\tau$-bench. Since the mid-training corpus contains only
Python code with no tool-use data, the cross-domain recovery is direct
evidence for the function-call/tool-call isomorphism.

\textbf{Contributions.}
\textbf{(1)} Framing function call sites as the internet-scale analogue of
the agent action--observation--continuation loop, motivating
function-granularity FIM as a self-supervised prior for agent capability.
\textbf{(2)} A function-aware FIM mid-training pipeline combining program
dependency graph analysis, a base-model-agnostic
complexity--inferability double criterion, and CoT rationales embedded
inside the FIM middle span.
\textbf{(3)} Robustness validation along three axes---two model sizes
(Qwen2.5-Coder-Instruct 7B/14B), two post-training pipelines (R2E-Gym, SWE-Smith),
and one alternative base (Qwen3-8B with SWE-Lego)---with consistent
in-domain gains in every configuration.
\textbf{(4)} A direct test of the motivating hypothesis: the same
Python-only corpus also yields gains on $\tau$-bench, BFCL, and
LiveCodeBench, evidence that the function-call inductive bias survives
post-training and transfers across task families.
\textbf{(5)} Open release of the 968-repository decontaminated corpus
($400\mathrm{K}$ FIM samples, $2.6\mathrm{B}$ tokens),
the selection pipeline, and mid-training checkpoints.

\section{Method}
\label{sec:method}
\vspace{-0.5ex}

\subsection{Motivation: Function Calls as Agent-Like Structures}
\label{sec:method:motivation}
A coding agent at step $t$ samples $a_t \sim \pi(a_t \mid h_t)$, observes
$o_{t+1} \sim p(o_{t+1} \mid h_t, a_t)$, and continues. Function calls
mirror this loop (Figure~\ref{fig:teaser}, left): pre-call context, call,
return, and downstream usage align with history, action, observation,
and continuation. This isomorphism motivates a fill-in-the-middle (FIM)
objective drawn from code. Random-span FIM~\citep{qwencoder,
deepseekcoder, starcoder} captures it only incidentally; our
\emph{function-aware} variant selects masking targets by program
structure and contextual predictability.



\vspace{-0.5ex}
\subsection{Data Collection and Decontamination}
\label{sec:method:data}

We curate a corpus of $968$ Python repositories from GitHub. Starting
from $\sim\!2{,}000$ candidates retrieved by combining a star-count
threshold with ten topic queries, we apply manual quality filtering and
remove every repository whose origin overlaps with the source
repositories of SWE-Bench~\citep{swebench} (verified by repository name
and any known fork). To eliminate test-time leakage, we restrict each
repository to commits whose timestamp precedes the earliest base-commit
used in SWE-Bench-Verified and SWE-Bench-Lite. The filtered corpus
yields $\approx\!400\mathrm{K}$ FIM samples ($\approx\!2.6\mathrm{B}$
tokens under the Qwen2.5-Coder tokenizer): $320\mathrm{K}$
single-function, $60\mathrm{K}$ pair, and $20\mathrm{K}$ triple targets.
Full statistics, topic-category breakdown, and license inventory are
reported in Appendix~\ref{app:data}.

\vspace{-0.5ex}
\subsection{Function-Aware FIM Target Selection}
\label{sec:method:selection}

\begin{figure}[t]
\centering
\includegraphics[width=\linewidth]{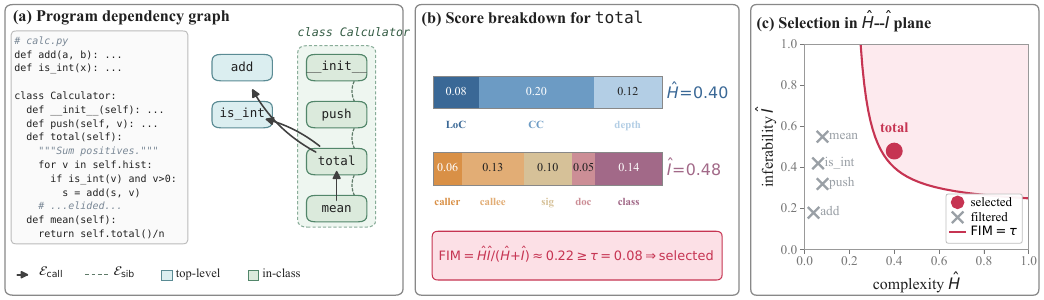}
\vspace{-2ex}
\caption{Function-aware FIM target selection on a small calculator
example. \textbf{(a)}~Program dependency graph parsed from the AST:
solid arrows are call edges $\mathcal{E}_{\mathrm{call}}$, dashed lines
are sibling edges $\mathcal{E}_{\mathrm{sib}}$ between same-class
methods. \textbf{(b)}~Stacked bars decompose the complexity score
$\hat{H}\!=\!0.40$ (Eq.~\ref{eq:complexity}; LoC, CC, depth) and the
inferability score $\hat{I}\!=\!0.48$ (Eq.~\ref{eq:inferability}; five
context signals) for $\mathtt{Calculator.total}$, yielding $\mathrm{FIM}\!\approx\!0.22\!\geq\!\tau\!=\!0.08$
(a hyperparameter; see Appendix~\ref{app:algo} for the full list).}
\label{fig:pdg_score}
\vspace{-1.5ex}
\end{figure}

Given each source file, our pipeline produces a set of mask targets
together with the corresponding masked file. Each target is a single
function or a connected group of $2$--$3$ functions; the multi-function
variant is studied in Section~\ref{sec:exp:ablation}. The pipeline has
four stages: dependency-graph construction, complexity scoring,
inferability scoring, and threshold-based selection. Figure~\ref{fig:pdg_score}
illustrates these stages on a small running example---a
\texttt{Calculator} class with two top-level helpers. The full
single-function selection algorithm is given in
Appendix~\ref{app:algo:single}, the multi-function extension in
Appendix~\ref{app:algo:groups}, and a numerical walkthrough of every
quantity shown in Figure~\ref{fig:pdg_score}(b) in
Appendix~\ref{app:algo:worked_example}.

\subsubsection{Program Dependency Graph}
\label{sec:method:pdg}

For each file we parse its AST and extract the set $\mathcal{V}$ of
function nodes (top-level functions and class methods, identified by
qualified names). Two edge sets are constructed: \emph{call edges}
$\mathcal{E}_{\mathrm{call}}$ between caller and callee, and
\emph{sibling edges} $\mathcal{E}_{\mathrm{sib}}$ between methods of the
same class (capturing intra-class coupling that flows through shared instance state rather than direct calls).
Call resolution handles common Python idioms (direct invocations, class
instantiation, \texttt{self}/\texttt{cls} method calls) with a
short-name fallback against the qualified-name index;
Appendix~\ref{app:algo:pdg} details the procedure.

\subsubsection{Complexity Score $\hat{H}$}
\label{sec:method:complexity}

For each $v \in \mathcal{V}$ we define
\vspace{-0.5ex}
\begin{equation}
  \hat{H}(v) \;=\; w_{\ell}\,\phi\!\left(\mathrm{LoC}(v),\,c_{\ell}\right)
            + w_{c}\,\phi\!\left(\mathrm{CC}(v),\,c_{c}\right)
            + w_{d}\,\phi\!\left(\mathrm{D}(v),\,c_{d}\right),
  \label{eq:complexity}
\end{equation}
where $\mathrm{LoC}(v)$ is lines of code, $\mathrm{CC}(v)$ is McCabe
cyclomatic complexity, $\mathrm{D}(v)$ is the maximum nesting depth of
control-flow constructs, and $\phi(x,c)\!=\!\min(x/c,\,2)$ normalizes
each quantity by a soft cap. Caps and weights are listed in
Appendix~\ref{app:algo:hyperparams}.

\subsubsection{Inferability Score $\hat{I}$}
\label{sec:method:inferability}

A target should be \emph{recoverable} from the surrounding context.
$\hat{I}(v)$ aggregates five context-derived signals that approximate
the mutual information between $v$'s body and the rest of the file:
\vspace{-0.5ex}
\begin{equation}
  \hat{I}(v) \;=\; \alpha\,C_{\mathrm{caller}}(v)
              + \beta\,C_{\mathrm{callee}}(v)
              + \gamma\,C_{\mathrm{sig}}(v)
              + \delta\,C_{\mathrm{doc}}(v)
              + \varepsilon\,C_{\mathrm{class}}(v).
  \label{eq:inferability}
\end{equation}
$C_{\mathrm{caller}}$ scores call-site argument specificity,
$C_{\mathrm{callee}}$ counts intra-file functions called by $v$,
$C_{\mathrm{sig}}$ aggregates type annotations and name descriptiveness,
$C_{\mathrm{doc}}$ indicates docstring presence, and
$C_{\mathrm{class}}$ counts in-class siblings; full component formulas
and weights are in Appendix~\ref{app:algo:inferability}. Each component
is a hand-designed proxy: a learned predictability score would couple
selection to a particular reference model and complicate the
cross-base-model generalization analysis (Section~\ref{sec:exp:main}).

\subsubsection{Single-Function Score}
\label{sec:method:single}

We combine $\hat{H}$ and $\hat{I}$ in a harmonic-mean-like form, scaled
by a one-sided difficulty penalty $\rho(\Delta(v))$:
\vspace{-0.5ex}
\begin{equation}
  \mathrm{FIM}(v) \;=\;
  \frac{\hat{H}(v)\,\hat{I}(v)}{\hat{H}(v) + \hat{I}(v) + \epsilon}
  \cdot \rho\!\left(\Delta(v)\right).
  \label{eq:fim_single}
\end{equation}
The harmonic-mean form forces both $\hat{H}$ and $\hat{I}$ to be large
simultaneously, penalizing imbalance; $\rho$ down-weights targets that
remain hard even given full context, which would otherwise be
unlearnable noise. Hard filters on length and dunder methods, the form
of $\rho$, and the threshold $\tau\!=\!0.08$ used throughout this work
are documented in Appendix~\ref{app:algo:penalty}.

\subsubsection{Multi-Function Group Selection}
\label{sec:method:groups}

Real-world code patches frequently span multiple related
functions~\citep{swebench}, motivating an extension that masks
\emph{groups} of $k\!=\!2$ or $k\!=\!3$ structurally connected functions;
we use this variant in our main recipe and ablate it in
Section~\ref{sec:exp:ablation}. A group score $\mathrm{FIM}(G)$
multiplies a coupling term, the harmonic-mean-like product of
group-level $\hat{H}(G)$ and $\hat{I}(G)$, and a difficulty penalty,
with $\hat{I}(G)$ recomputed under joint masking so that intra-group
references cannot inflate the score. The eight topology patterns
(caller-callee, co-callee, sibling-coupled, mutual-call, call-chain,
hub, fan-in, class-triad), the full equations,
Algorithm~\ref{alg:selection_multi}, and a worked pair example are in
Appendix~\ref{app:algo:groups}.

\vspace{-0.5ex}
\subsection{Chain-of-Thought Augmentation}
\label{sec:method:cot}
For each selected target we run a three-stage pipeline.
\textbf{Generate.} Gemini-3-Flash sees only the masked file and
produces a step-by-step rationale together with a candidate function
body, with no access to the ground-truth body
(Appendix~\ref{app:algo:cot-gen}).
\textbf{Filter.} A separate Gemini-3-Flash judge scores the
(rationale, candidate body) pair against the ground-truth body on
feasibility and five quality dimensions; we keep the top-scoring
$\sim\!400$K samples (Appendix~\ref{app:algo:filter}).
\textbf{Format.} Each retained pair is placed inside the FIM middle
span, rationale before body (Appendix~\ref{app:algo:cot-format}):
\vspace{-0.5ex}
\begin{center}\small
\texttt{<fim\_prefix>}\,$\langle$prefix$\rangle$\,%
\texttt{<fim\_suffix>}\,$\langle$suffix$\rangle$\,%
\texttt{<fim\_middle>}\,$\langle$rationale$\rangle$\,$\langle$body$\rangle$
\end{center}
\vspace{-1ex}
The model is thus trained to produce reasoning followed by consistent
code, mirroring the think-then-act structure of an agent step. The
ground-truth body serves only as a filter anchor and never appears in
the training target.
Section~\ref{sec:exp:ablation} isolates the contribution of FIM
structure from CoT distillation via a \emph{self-CoT} variant in
which the model under training generates its own rationales.


\section{Experiments}
\label{sec:exp}
\vspace{-0.5ex}

\subsection{Setup}
\label{sec:exp:setup}

\textbf{Benchmarks.} We evaluate on three groups.
\emph{Coding-agent benchmarks}: SWE-Bench-Verified ($500$) and
SWE-Bench-Lite ($300$)~\citep{swebench}, our primary in-domain target.
\emph{Non-agent coding benchmarks}: LiveCodeBench~\citep{livecodebench},
OJBench~\citep{ojbench}, and FullStackBench-EN~\citep{fullstackbench};
these probe pure code generation and serve as a regression check.
\emph{Tool-use and OOD agent benchmarks}: Terminal-Bench 2.0~\citep{terminalbench},
$\tau$-bench~\citep{taubench}, and BFCL~\citep{bfcl}; the latter two
contain no Python code-editing trajectories and test whether the
function-call inductive bias transfers across task families.

\textbf{Mid-training and post-training pipeline.}
We mid-train Qwen2.5-Coder-Instruct (7B/14B)~\citep{qwencoder} and
Qwen3-8B~\citep{qwen3} on the selected FIM corpus
using the standard FIM loss on the middle span
(rationale plus body), packed to the model's native context length and
using its native FIM sentinel tokens. We then apply existing agentic
post-training pipelines without modification:
R2E-Gym~\citep{r2egym} or SWE-Smith~\citep{swesmith} for the
Qwen2.5-Coder runs, and SWE-Lego~\citep{swelego} for Qwen3-8B. For
Qwen3-8B we train SWE-Lego for $2$ epochs rather than the official $4$
to prevent overfitting in our setup. Hyperparameters and token budgets
are in Appendix~\ref{app:hp}.

\textbf{Evaluation protocol.}
All numbers are means over three independent evaluation seeds on the
final checkpoint of each pipeline, with ``\%'' omitted in table cells
and std bands in parentheses. \emph{Baseline} is base $+$ post-training;
\emph{ours} is base $+$ FIM mid-training $+$ identical post-training.
We do not evaluate mid-training-only checkpoints because FIM-only models
have degraded instruction-following and cannot be compared fairly with
instruction-tuned baselines---every reported gain therefore
\emph{survives} subsequent post-training. The agent harness is fixed by
the post-training pipeline: R2E-Gym uses an OpenHands fork~\citep{openhands,
r2egym}, SWE-Smith uses SWE-agent~\citep{sweagent}, and the Qwen3-8B
runs use OpenHands directly (SWE-Lego data exceeds the 32K context of
the Qwen2.5-Coder models).

\textbf{Models and baselines.}
For Qwen2.5-Coder-Instruct at 7B/14B~\citep{qwencoder} we report
(i) the instruction-tuned base, (ii) post-training with R2E-Gym
reproduced under our setup, (iii) the official R2E-Gym numbers (in grey,
where available), (iv) our recipe of FIM mid-training followed by
R2E-Gym post-training, and at 7B additionally the analogous
SWE-Smith~\citep{swesmith} variants. Qwen3-8B uses the
SWE-Lego~\citep{swelego} pipeline; the cross-base-model comparison thus
varies the post-training pipeline simultaneously, a confound we discuss
in Section~\ref{sec:exp:main}.

\vspace{-0.5ex}
\subsection{Main Results on SWE Agent Benchmarks}
\label{sec:exp:main}

\begin{table}[t]
\centering
\caption{Main results on coding agent benchmarks. All numbers are
percentages, means over three independent seeds with std bands;
\emph{Average} is the unweighted mean of Verified and Lite. Bolded rows
are our recipe; rows tagged \emph{officially reported} are quoted from
the corresponding original publications. SWE-Lego is post-trained for
$2$ epochs (vs.\ $4$ in the official release) to prevent overfitting.}
\label{tab:main}
\setlength{\tabcolsep}{6pt}
\begin{tabular}{lccc}
\toprule
Setting & SWE-Bench-Verified & SWE-Bench-Lite & Average \\
\midrule
\multicolumn{4}{l}{\textit{Qwen2.5-Coder-7B-Instruct}} \\
\quad --- (no agentic training)                       & 1.80\,($\pm$1.30)            & 1.00\,($\pm$1.00)            & 1.40 \\
\quad + R2E-Gym~\citep{r2egym} (reproduced)           & 15.00\,($\pm$1.50)           & 11.33\,($\pm$1.20)           & 13.17 \\
\quad + R2E-Gym (officially reported)                 & 19.00\,($\pm$1.00)           & 11.00\,($\pm$0.80)           & 15.00 \\
\quad \textbf{+ FIM-Midtrain + R2E-Gym}               & \textbf{17.80}\,($\pm$1.40)  & \textbf{15.00}\,($\pm$1.10)  & \textbf{16.40} \\
\rowcolor{blue!8}
\quad\quad $\Delta$ (ours vs.\ reproduced)            & $+2.80$                      & $+3.67$                      & $+3.24$ \\
\cmidrule(lr){1-4}
\quad + SWE-Smith~\citep{swesmith} (reproduced)       & 12.30\,($\pm$1.20)           & 14.20\,($\pm$1.40)           & 13.25 \\
\quad + SWE-Smith (officially reported)               & 15.20                        & 11.70                        & 13.45 \\
\quad \textbf{+ FIM-Midtrain + SWE-Smith}             & \textbf{17.60}\,($\pm$1.30)  & \textbf{14.70}\,($\pm$1.00)  & \textbf{16.15} \\
\rowcolor{blue!8}
\quad\quad $\Delta$ (ours vs.\ reproduced)            & $+5.30$                      & $+0.50$                      & $+2.90$ \\
\midrule
\multicolumn{4}{l}{\textit{Qwen2.5-Coder-14B-Instruct}} \\
\quad --- (no agentic training)                       & 4.00\,($\pm$1.60)            & 2.70\,($\pm$1.00)            & 3.35 \\
\quad + R2E-Gym (reproduced)                          & 26.20\,($\pm$1.40)           & 18.00\,($\pm$1.10)           & 22.10 \\
\quad + R2E-Gym (officially reported)                 & 26.80\,($\pm$1.40)           & 20.67\,($\pm$0.70)           & 23.74 \\
\quad \textbf{+ FIM-Midtrain + R2E-Gym}               & \textbf{29.20}\,($\pm$1.50)  & \textbf{22.00}\,($\pm$1.20)  & \textbf{25.60} \\
\rowcolor{blue!8}
\quad\quad $\Delta$ (ours vs.\ reproduced)            & $+3.00$                      & $+4.00$                      & $+3.50$ \\
\midrule
\multicolumn{4}{l}{\textit{Qwen3-8B}} \\
\quad --- (no agentic training)                       & 7.60\,($\pm$1.20)            & 5.80\,($\pm$0.90)            & 6.70 \\
\quad + SWE-Lego~\citep{swelego} (reproduced)         & 31.80\,($\pm$1.00)           & 27.30\,($\pm$1.10)           & 29.55 \\
\quad \textbf{+ FIM-Midtrain + SWE-Lego}              & \textbf{35.00}\,($\pm$1.50)  & \textbf{32.70}\,($\pm$1.30)  & \textbf{33.85} \\
\rowcolor{blue!8}
\quad\quad $\Delta$ (ours vs.\ reproduced)            & $+3.20$                      & $+5.40$                      & $+4.30$ \\
\bottomrule
\end{tabular}
\end{table}

Table~\ref{tab:main} summarizes in-domain agent results.

\textbf{Consistent gains on the Qwen2.5-Coder-Instruct series.}
Holding the post-training pipeline fixed at R2E-Gym, FIM mid-training
improves SWE-Bench-Verified by $+2.80$ on 7B-Instruct and $+3.00$ on
14B-Instruct, with matching directional gains on SWE-Bench-Lite
($+3.67/+4.00$). Both Qwen2.5-Coder-Instruct sizes benefit from the same
mid-training corpus and recipe, indicating that the structural prior is
not absorbed by the larger pretrained checkpoint within this family.

\textbf{Transfer across post-training pipelines.}
Replacing R2E-Gym with SWE-Smith on the same 7B base yields $+5.30$
points on Verified, larger than the $+2.80$ under R2E-Gym; on Lite the
SWE-Smith pairing gains only $+0.50$, smaller than $+3.67$ for the
R2E-Gym pairing. The two pipelines together indicate that mid-training
is not tuned to a single post-training data distribution, though the
magnitude of its benefit depends on the baseline pipeline it composes
with.

\textbf{Transfer to a non-Qwen2.5 base.}
Switching to Qwen3-8B paired with SWE-Lego, mid-training improves
Verified by $+3.20$ and Lite by $+5.40$, comparable to the Qwen2.5-Coder-Instruct
gains. This single comparison varies the post-training pipeline jointly
with the base model, so the
result should be read as ``not specific to the Qwen2.5-Coder-Instruct $+$
R2E-Gym/SWE-Smith pairing'' rather than as a guarantee across
base-model families.

\vspace{-0.5ex}
\subsection{Capability Preservation and Cross-Domain Transfer}
\label{sec:exp:gen}

A natural concern with task-specialized post-training is that it erodes
capabilities the base model already had. We further evaluate the 14B model on six additional benchmarks to assess
capability preservation. To save compute, we restrict this controlled
comparison (instruct base vs.\ post-training only vs.\ mid-training $+$
post-training) to R2E-Gym as the post-training dataset (Table~\ref{tab:cap}).

\begin{table}[t]
\centering
\caption{Capability preservation and cross-domain transfer at 14B with
R2E-Gym. Bold marks the better trained variant per column (post-only
vs.\ ours); the Instruct row is shown as a reference ceiling. All cells
are percentages. ``Terminal'' denotes Terminal-Bench 2.0.}
\label{tab:cap}
\small
\setlength{\tabcolsep}{4pt}
\begin{tabular}{lcccccc|c}
\toprule
& \multicolumn{3}{c}{Non-agent coding}
& \multicolumn{1}{c}{Agent OOD}
& \multicolumn{2}{c}{Tool use}
& \\
\cmidrule(lr){2-4} \cmidrule(lr){5-5} \cmidrule(lr){6-7}
Setting
& LiveCode & OJBench & FSB-EN
& Terminal
& $\tau$-bench & BFCL
& Avg \\
\midrule
Instruct Model
  & 37.20 & 5.20 & 53.80
  & 0.00
  & 5.70 & 23.20
  & 20.85 \\
+ R2E-Gym
  & 24.10 & 2.80 & 47.72
  & 2.41
  & 3.40 & 15.80
  & 16.04 \\
\textbf{+ FIM Mid-Train + R2E-Gym}
  & \textbf{35.20} & \textbf{4.74} & \textbf{48.25}
  & \textbf{3.66}
  & \textbf{7.30} & \textbf{18.20}
  & \textbf{19.56} \\
\rowcolor{blue!8}
$\Delta$ (vs.\ R2E-Gym only)
  & $+11.10$ & $+1.94$ & $+0.53$
  & $+1.25$
  & $+3.90$ & $+2.40$
  & $+3.52$ \\
\bottomrule
\end{tabular}
\end{table}

\textbf{Agentic post-training has a substantial hidden capability cost.}
R2E-Gym alone reduces every non-agent and tool-use benchmark relative to
the Instruct ceiling: LiveCodeBench by $13.10$, BFCL by $7.40$,
FullStackBench-EN by $6.08$, $\tau$-bench by $2.30$, and OJBench by
$2.40$. Averaged across the six benchmarks the post-trained model loses
$4.81$ points relative to Instruct---the implicit cost paid for
SWE-Bench gains, rarely highlighted in agent papers.

\textbf{Mid-training largely closes the regression gap.}
Adding FIM mid-training before the same post-training restores
LiveCodeBench by $+11.10$, OJBench by $+1.94$ (within $0.46$ of the
Instruct ceiling), and FullStackBench-EN by $+0.53$. The six-benchmark
average rises from $16.04$ to $19.56$ ($+3.52$ over post-training only),
while the SWE-Bench-Verified gain on the same base is preserved
(Table~\ref{tab:main}). Mid-training therefore improves the
cost--benefit profile of agentic post-training: in-domain target
metrics improve and the bulk of off-distribution erosion is undone in
the same training run.

\textbf{Cross-domain transfer to non-coding tool use.}
$\tau$-bench and BFCL contain no Python code-editing data, and our
mid-training corpus contains no tool-use trajectories. Mid-training
nevertheless improves $\tau$-bench by $+3.90$ and BFCL by $+2.40$ over
post-training alone, with a consistent recovery on
Terminal-Bench 2.0 ($+1.25$). Because the corpus carries no tool-use signal, the only mechanism is a
structural prior installed at mid-training that survives post-training, which is the direct evidence for the function-call/tool-call isomorphism(Section~\ref{sec:method:motivation}).

\vspace{-0.5ex}
\subsection{Ablation Studies}
\label{sec:exp:ablation}

We run three controlled ablations on Qwen2.5-Coder-7B-Instruct with
R2E-Gym post-training and a shared mid-training-free baseline. The first
isolates the role of the chain-of-thought rationale; the second isolates
the role of the function-aware selection pipeline; the third varies
mask granularity (single-function vs.\ multi-function groups). Due to
compute constraints, ablations run at 7B only.

\begin{table}[t]
\centering
\caption{Ablations on the 7B model with R2E-Gym post-training.
\textbf{(A)}~rationale source. \textbf{(B)}~function-selection
algorithm. \textbf{(C)}~mask granularity (single vs.\ multi-function
groups). All blocks share the baseline (\emph{w/o mid-train}) and a
controlled $200$K-target budget; CoT is fixed to Gemini-3-Flash in
(B) and (C). Bold marks the best configuration per block; the
$80\%/15\%/5\%$ mixture in~(C) is the recipe used in our main results
(Table~\ref{tab:main}), which is trained on the full corpus rather
than the $200$K budget here. Absolute numbers across all blocks
therefore lie below the main-table recipe; the relative orderings
within each block, rather than the absolute values, are the object
of comparison.}
\label{tab:ablation}
\small
\setlength{\tabcolsep}{8pt}
\begin{tabular}{lccc}
\toprule
Setting & SWE-Bench-Verified & SWE-Bench-Lite & Average \\
\midrule
\multicolumn{4}{l}{\emph{(A) Chain-of-thought source (selection fixed to ``full'')}} \\
\midrule
w/o mid-train (baseline)                  & 15.00 & 11.33 & 13.17 \\
+ FIM, no CoT                             & 16.10 & 12.60 & 14.35 \\
+ FIM, self-CoT (Qwen2.5-Coder-7B-Instruct) & 16.40 & 13.30 & 14.85 \\
\textbf{+ FIM, Gemini-3-Flash CoT}        & \textbf{17.00} & \textbf{14.20} & \textbf{15.60} \\
\midrule
\multicolumn{4}{l}{\emph{(B) Function-selection algorithm (CoT fixed to Gemini-3-Flash)}} \\
\midrule
w/o mid-train (baseline)                    & 15.00 & 11.33 & 13.17 \\
Random                                      & 15.30 & 12.60 & 13.95 \\
Gemini-selected                             & 16.40 & 13.70 & 15.05 \\
PDG only                                    & 16.10 & 13.60 & 14.85 \\
PDG + complexity ($\hat{H}$)                & 16.50 & 13.60 & 15.05 \\
PDG + inferability ($\hat{I}$)              & 16.70 & 14.00 & 15.35 \\
\textbf{Full (PDG + $\hat{H}$ + $\hat{I}$)} & \textbf{17.00} & \textbf{14.20} & \textbf{15.60} \\
\midrule
\multicolumn{4}{l}{\emph{(C) Mask granularity (selection fixed to ``full'', CoT fixed to Gemini-3-Flash)}} \\
\midrule
w/o mid-train (baseline)             & 15.00 & 11.33 & 13.17 \\
Single-function only                 & 17.00 & 14.20 & 15.60 \\
85\% single + 15\% pair ($k=2$)       & 17.20 & 14.60 & 15.90 \\
95\% single + 5\% triple ($k=3$)      & 17.00 & 14.40 & 15.70 \\
\textbf{80\% single + 15\% pair + 5\% triple} & \textbf{17.40} & \textbf{14.80} & \textbf{16.10} \\
\bottomrule
\end{tabular}
\end{table}

\textbf{FIM structure contributes independently of CoT distillation.}
Block~(A) addresses the concern that gains stem mainly from distilling
a frontier teacher. Removing the rationale entirely (\emph{no CoT})
already lifts the average by $+1.18$---roughly half of the $+2.43$-point
Gemini-3 gain---direct evidence that the function-aware FIM signal does
substantive work before any reasoning supervision is added. Replacing
Gemini-3 with rationales from the model under training
(\emph{self-CoT}) reaches $14.85$, recovering $1.68$ of the $2.43$-point
gap; the residual $0.75$ points attributable to a frontier teacher are
real but modest. The recipe is therefore not a thinly disguised
distillation pipeline.

\textbf{Function-selection algorithm is the dominant lever.}
Block~(B) varies the selection algorithm with CoT and budget held
fixed. \emph{Random} masking sets a floor at $13.95$, and
\emph{Gemini-selected} reaches $15.05$: frontier judgment on
\emph{which} function to mask helps but is not sufficient. Restricting
candidates to functions with at least one PDG neighbor (\emph{PDG only})
reaches $14.85$; adding $\hat{H}$ or $\hat{I}$ on top yields $15.05$ and
$15.35$---intrinsic difficulty and contextual recoverability each
contribute beyond the structural filter. Combining them (\emph{Full})
reaches $15.60$, confirming that $\hat{H}$ and $\hat{I}$ are not
redundant.

\textbf{Mask granularity: pair masking helps, triples saturate.}
Block~(C) mixes single-function targets with multi-function groups
(Section~\ref{sec:method:groups}). Adding $15\%$ pair targets raises
the average from $15.60$ to $15.90$; substituting $5\%$ triple targets
instead is essentially neutral ($15.70$). The full $80\%/15\%/5\%$ mix
achieves $16.10$. This matches the analysis in
Section~\ref{sec:analysis:multifn}: training on cross-function
dependencies disproportionately helps tasks whose gold patches span
multiple functions, but the marginal return from larger groups
diminishes as coupling becomes harder to maintain under joint masking.


\section{Analysis}
\label{sec:analysis}
\vspace{-0.5ex}

The previous section established that FIM mid-training improves end-task
metrics. We now ask \emph{how} the resulting agent behaves differently
and \emph{where} along the trajectory the gains accrue. Both analyses
use the 14B configuration with R2E-Gym, comparing the
post-training-only baseline (\emph{R2E-Gym}) against our recipe
(\emph{FIM-Midtrain $+$ R2E-Gym}). The Lite analogue, the full
action-type distribution, the no-patch mechanism, and a concrete
trajectory contrast are deferred to Appendix~\ref{app:behavior_extra}.

\vspace{-0.5ex}
\subsection{Recovery from Negative Observations}
\label{sec:analysis:recovery}

A trajectory contains a \emph{negative observation} if any tool output
matches a fixed set of error patterns (stack traces,
``\texttt{No replacement was performed}'', shell errors, etc.; full list
in Appendix~\ref{app:algo}). The fraction of such trajectories is
essentially identical across checkpoints ($88.8\%$ baseline vs.\
$91.8\%$ ours), so the agents see comparable amounts of negative
feedback. The \emph{recovery rate}---the fraction of error-containing
trajectories that nonetheless terminate with a passing patch---rises
from $24.8\%$ to $28.8\%$ ($+4.0$~pp; $+16\%$ relative;
Table~\ref{tab:behavior_main}). This is precisely the capability our
framing predicts mid-training should support: continuing productively
after an external return contradicts the model's prior expectation.
Mid-training also shifts the agent toward an iterate-and-verify policy,
increasing edits per solved task from $3.3$ to $7.4$ and trajectory
length from $15.1$ to $23.6$ steps; action-type breakdowns are in
Appendix~\ref{app:behavior_extra}.

\begin{table}[h]
\centering
\caption{Headline trajectory metrics on SWE-Bench-Verified. Full metrics in Appendix~\ref{app:behavior_extra}.}
\label{tab:behavior_main}
\small
\setlength{\tabcolsep}{8pt}
\vspace{-0.5ex}
\begin{tabular}{lcccc}
\toprule
Setting & Recovery rate (\%) & Edits / solved & Steps / solved & Pass (\%) \\
\midrule
+ R2E-Gym                          & 24.8           & 3.3           & 15.1           & 26.2 \\
\textbf{+ FIM-Midtrain + R2E-Gym}  & \textbf{28.8}  & \textbf{7.4}  & \textbf{23.6}  & \textbf{29.2} \\
\bottomrule
\end{tabular}
\vspace{-1ex}
\end{table}

\vspace{-0.5ex}
\subsection{The Gain Concentrates on Multi-Function Reasoning}
\label{sec:analysis:multifn}

The most direct test of the isomorphism argument
(Section~\ref{sec:method:motivation}) is whether mid-training
preferentially helps tasks that require reasoning about cross-function
dependencies inside a file. We stratify the $500$ Verified tasks by the
shape of the gold patch (Figure~\ref{fig:patchshape},
Appendix~\ref{app:behavior_extra}). On the $88$ tasks whose gold patch
modifies $\geq 2$ functions within a single file, the baseline solves
$13.6\%$ on average and ours solves $22.7\%$, an absolute gain of
$+9.1$~pp---more than $4{\times}$ the gain on the $341$ single-function
tasks ($+2.1$~pp). Per-instance head-to-head on this bucket shows ours
uniquely solves about twice as many tasks as the baseline. Multi-file
tasks ($n{=}71$) are not differentially helped, which we attribute to a
granularity mismatch: our FIM operates within files
(Appendix~\ref{app:behavior_extra}). The slice where mid-training helps
most is precisely the slice where the agent must follow control- and
data-dependencies between caller-callee or sibling functions inside a
file---the same structure exposed by function-aware FIM masking. A
complementary outcome-distribution breakdown
(Figure~\ref{fig:failure_app}) shows the $+15$-task gain is driven by an
order-of-magnitude reduction in \emph{no-patch} failures alongside small
reductions in localization errors; the FIM signal conditions the model
to produce a non-empty span between prefix and suffix---a disposition
that survives subsequent post-training.

\section{Related Work}
\vspace{-0.5ex}
\label{sec:related}

\textbf{Mid-training and continued pretraining.}
A growing body of work~\citep{mid-training-survey, gururangan2020dontstop}
identifies a stage between pretraining and post-training in which a model
is further trained on a curated corpus to install inductive biases hard
to acquire from generic web text or fine-tuning. MiniCPM~\citep{minicpm},
OLMo~\citep{olmo}, and DeepSeek-V3~\citep{deepseekv3} schedule such a
stage near the end of pretraining; Code-Llama~\citep{codellama} follows
the same staging philosophy for context-length generalization. Recent
work shows \emph{when} specialized data is introduced matters as much as
how much is used~\citep{remit, frontload-reasoning}. We adopt this
philosophy but target an \emph{agent-oriented} structural prior through
function-aware FIM.

\textbf{Fill-in-the-middle and structure-aware code objectives.}
FIM~\citep{bavarian2022fim, fried2022incoder} is a defining ingredient
of modern code LLM pretraining, used in
Code-Llama~\citep{codellama}, StarCoder-2~\citep{starcoder2},
CodeGen~\citep{codegen}, Qwen-Coder~\citep{qwencoder}, and
DeepSeek-Coder~\citep{deepseekcoder}.
A recent line moves beyond random spans toward \emph{structure-aware}
masking: AST-T5~\citep{ast-t5} and AST-FIM~\citep{ast-fim} mask AST
subtrees, the latter reporting up to $+5$ pts on real-edit FIM
benchmarks; Horizon-Length Prediction~\citep{hlp} adds a planning signal;
Instruction-aware FIM~\citep{ifim} extends the FIM tuple with a
developer-comment slot. Repository-level retrieval methods such as
GraphCoder~\citep{graphcoder} and DRACO~\citep{draco} exploit program
dependencies, but at inference time. Our recipe differs along three
axes: targets are chosen at \emph{function granularity} via PDG analysis
with a complexity--inferability criterion, so the masked region is a
unit of agent-relevant reasoning rather than a syntactic subtree; an
explicit \emph{chain-of-thought} sits inside the FIM middle; and the
objective lives in a \emph{dedicated mid-training stage}.

\textbf{Coding agent foundation models.}
Progress has advanced on three fronts. Benchmarks have moved from
SWE-Bench~\citep{swebench} to broader, contamination-resistant suites
including Multi-SWE-Bench~\citep{multiswebench} and
SWE-Bench-Pro~\citep{swebenchpro}. Scaffolds such as
SWE-agent~\citep{sweagent}, OpenHands~\citep{openhands}, and
Agentless~\citep{agentless} expose different action interfaces.
Trajectory-centric post-training pipelines---SWE-Gym~\citep{swegym},
R2E-Gym~\citep{r2egym}, SWE-Smith~\citep{swesmith},
SWE-Lego~\citep{swelego}, and Skywork-SWE~\citep{skyworkswe}---curate
agent trajectories on top of these scaffolds, with recent extensions
replacing or augmenting SFT with RL~\citep{longcontext-swerl,
selfplay-swerl}. We use R2E-Gym, SWE-Smith, and SWE-Lego unmodified
atop a mid-trained checkpoint; the novelty lies one stage earlier---a
self-supervised signal extracted from structure already present in
source code, which mitigates the off-domain capability erosion
trajectory-only post-training inflicts.

\textbf{Distillation from frontier models.}
Our recipe uses Gemini-3-Flash to generate the CoT rationales embedded
in each FIM middle, placing it within distillation from a strong
teacher~\citep{orca, distillwhisper, magicoder} and CoT
distillation~\citep{ho2023reasoningteachers, hsieh2023distill,
chen2025cotdistill}. Recent work shows diversity and structural
alignment of the rationale matter as much as teacher
strength~\citep{chen2025cotdistill}; consistent with this, our
CoT-source ablation (Section~\ref{sec:exp:ablation}) confirms the
recipe is not distillation-bound.

\section{Limitations and Discussion}
\vspace{-0.5ex}
\label{sec:limitations}
We close by stating four limitations that scope our claims.
(i) \textit{Python-only corpus and evaluation.} The mid-training corpus
and the in-domain agent benchmarks are exclusively Python; cross-language
evidence comes only indirectly through FullStackBench-EN
(Section~\ref{sec:exp:gen}), and transfer to Java, C++, or Rust is left
to future work.
(ii) \textit{Teacher dependency for CoT.} The default recipe relies on
Gemini-3-Flash; the CoT-source ablation (Section~\ref{sec:exp:ablation})
shows self-generated rationales recover most of the gain, but a fully
open-source replication requires a comparably strong open teacher.
(iii) \textit{Partial cross-base validation.} Our cross-base evidence
comes from a single non-Qwen2.5-Coder configuration (Qwen3-8B with
SWE-Lego), which simultaneously varies the post-training pipeline; the
result indicates the recipe is not tied to one pretraining/post-training
combination rather than guaranteeing transfer across all base families.
(iv) \textit{Modularity assumption.} Function-aware FIM presupposes
modular code; on monolithic scripts, generated code, or notebooks, the
pipeline yields fewer eligible targets, a regime we do not study
systematically.

\section{Conclusion}
\vspace{-0.5ex}
\label{sec:conclusion}
A single step of a coding agent and a single function call site share
the same four-part structure---\textit{context, action, externally
produced return, continuation}---making source code an internet-scale
supply of agent-relevant signal. We turn this into function-aware FIM
mid-training: a self-supervised stage that masks targets selected via
program dependency graph analysis and a complexity--inferability
double criterion, with chain-of-thought rationales embedded inside the
FIM middle. Across model size, post-training pipeline, and base-model
family, the recipe delivers consistent gains on coding-agent
benchmarks, and the same Python-only corpus transfers to non-coding
tool-use benchmarks ($\tau$-bench, BFCL). Future work includes
extending the selection to non-Python languages and composing
mid-training with RL post-training.

\bibliography{references}
\bibliographystyle{plain}


\appendix


\section{Corpus Details}
\label{app:data}

This appendix expands on the data-collection summary
(Section~\ref{sec:method:data}). We report category coverage, per-property
statistics, and the license inventory of the $968$ released repositories.

\begin{figure}[h]
\centering
\includegraphics[width=0.65\linewidth]{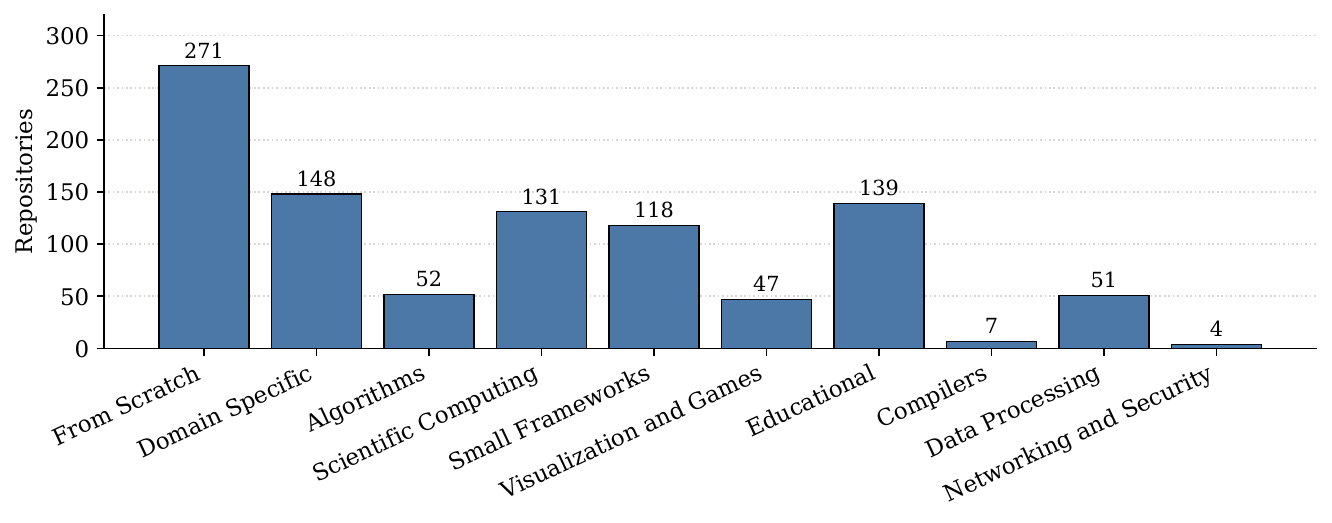}
\caption{Distribution of the $968$ source repositories across ten topic
categories. The corpus is dominated by reference implementations,
scientific computing, and small frameworks; compiler and
networking/security tails are kept by design to maintain coverage
diversity.}
\label{fig:dataset_categories}
\end{figure}

\begin{table}[h]
\centering
\caption{Mid-training corpus statistics after decontamination and
quality filtering. Token counts use the Qwen2.5-Coder tokenizer.
All numeric quantities reflect the released corpus.}
\label{tab:dataset}
\small
\setlength{\tabcolsep}{6pt}
\begin{tabular}{lr}
\toprule
Property & Value \\
\midrule
Source GitHub repositories                       & $968$ \\
Topic categories                                 & $10$ \\
Filtered self-contained Python files             & $\approx 78\mathrm{K}$ \\
Total FIM samples                                & $\approx 400\mathrm{K}$ \\
Mid-training token budget                        & $\approx 2.6\mathrm{B}$ \\
\midrule
Single-function FIM targets                      & $\approx 320\mathrm{K}$ ($\approx 2.0\mathrm{B}$ tokens) \\
Multi-function targets ($k{=}2$)                 & $\approx 60\mathrm{K}$ ($\approx 0.4\mathrm{B}$ tokens) \\
Multi-function targets ($k{=}3$)                 & $\approx 20\mathrm{K}$ ($\approx 0.2\mathrm{B}$ tokens) \\
Mean target LoC                                  & $\approx 34$ \\
Targets with Gemini-3 CoT                        & $100\%$ \\
\midrule
SWE-Bench source-repo overlap                    & $0$ \\
\bottomrule
\end{tabular}
\end{table}

\subsection{License Inventory}
\label{app:data:license}

\begin{figure}[h]
\centering
\includegraphics[width=0.85\linewidth]{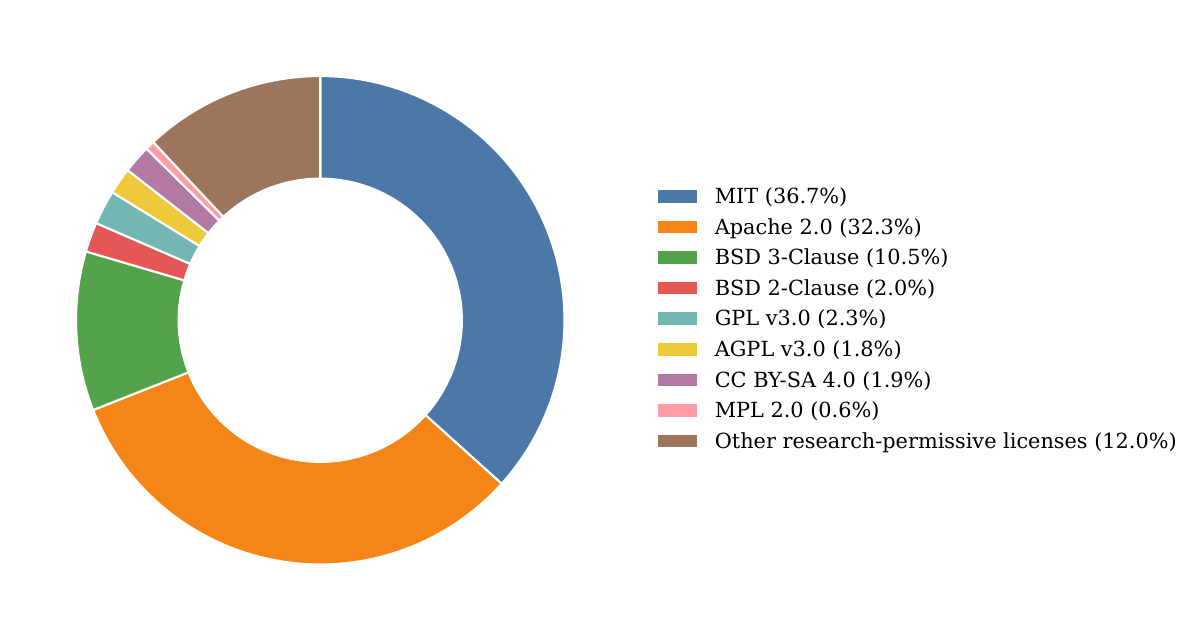}
\caption{License distribution of the $968$-repository corpus. Permissive
licenses (MIT, Apache-2.0, BSD) account for over $80\%$ of the corpus.
Small categories (LGPL, ISC, Boost, MIT-0, Unlicense, CC0, CC BY,
CC BY-NC, etc.) are aggregated as ``Other research-permissive licenses,''
all of which permit at least non-commercial research use.}
\label{fig:licenses}
\end{figure}

Figure~\ref{fig:licenses} shows the license distribution across the
$968$ released repositories. The corpus is dominated by permissive
licenses (MIT, Apache-2.0, BSD-3); the remainder is split among
copyleft (GPL, AGPL, LGPL, MPL) and Creative Commons families. Every
license in the corpus permits at least non-commercial research use, and
all small categories are aggregated in this chart as
``Other research-permissive licenses.'' The per-repository license is
released alongside the repository list.

\section{Algorithmic Details}
\label{app:algo}

This appendix collects the full algorithmic details summarized in
Section~\ref{sec:method:selection}. The complete Python implementation
of both the single-function and multi-function selection pipelines,
including all default hyperparameters listed below, is provided in the
supplementary material.

\subsection{Single-Function FIM Target Selection}
\label{app:algo:single}

Algorithm~\ref{alg:selection} summarizes the per-file pipeline that
produces single-function FIM targets. It composes the program
dependency graph (Section~\ref{sec:method:pdg}), the complexity score
$\hat{H}$ (Section~\ref{sec:method:complexity}), the inferability score
$\hat{I}$ (Section~\ref{sec:method:inferability}), the difficulty
penalty $\rho$ (Appendix~\ref{app:algo:penalty}), and the hard filters
described below.

\begin{algorithm}[h]
\caption{Single-function FIM target selection for one file.}
\label{alg:selection}
\begin{algorithmic}[1]
  \Require Source file $s$; thresholds $\tau_d, \tau_{\mathrm{FIM}}, \tau_{\hat H}$;
           file-line bounds $[L_{\min}, L_{\max}]$; LoC bounds $[\ell_{\min}, \ell_{\max}]$
  \If{$\mathrm{Lines}(s) \notin [L_{\min}, L_{\max}]$}
     \State \Return $\emptyset$ \Comment{file-level pre-filter}
  \EndIf
  \State $T \gets \mathrm{ParseAST}(s)$
  \State $\mathcal{V}, \mathcal{E}_{\mathrm{call}}, \mathcal{E}_{\mathrm{sib}} \gets \mathrm{BuildPDG}(T)$
        \Comment{Section~\ref{sec:method:pdg}}
  \State $\mathcal{C} \gets \emptyset$
  \For{each $v \in \mathcal{V}$}
    \If{$v$ is a dunder method \textbf{or} $\mathrm{LoC}(v) \notin [\ell_{\min}, \ell_{\max}]$}
      \State \textbf{continue}
    \EndIf
    \State Compute $\hat{H}(v)$ and $\hat{I}(v)$
          \Comment{Eqs.~\eqref{eq:complexity}, \eqref{eq:inferability}}
    \If{$\hat{H}(v) < \tau_{\hat H}$} \textbf{continue} \EndIf
    \State $\Delta(v) \gets \max(0, \hat{H}(v)-\hat{I}(v)) / (\hat{H}(v)+\epsilon)$
    \State $\mathrm{FIM}(v) \gets \dfrac{\hat{H}(v)\,\hat{I}(v)}{\hat{H}(v)+\hat{I}(v)+\epsilon}\cdot \rho(\Delta(v))$
    \If{$\mathrm{FIM}(v) \ge \tau_{\mathrm{FIM}}$}
      \State $\mathcal{C} \gets \mathcal{C} \cup \{v\}$
    \EndIf
  \EndFor
  \State \Return $\mathcal{C}$ \Comment{ranked by $\mathrm{FIM}$ descending}
\end{algorithmic}
\end{algorithm}

We use $L_{\min}=50$, $L_{\max}=1800$, $\ell_{\min}=10$, $\ell_{\max}=200$,
$\tau_{\hat H}=0.15$, and $\tau_{\mathrm{FIM}}=0.08$ throughout.

\subsection{Program Dependency Graph: Call Resolution}
\label{app:algo:pdg}

For every \texttt{Call} node inside the body of $u \in \mathcal{V}$, we
attempt to resolve its callee to some $v \in \mathcal{V}$, adding a
directed edge $u \to v$ in $\mathcal{E}_{\mathrm{call}}$. Resolution
handles three Python idioms:
\begin{itemize}\setlength\itemsep{0.1em}
  \item \textbf{Direct module-level invocations}: \texttt{foo(...)} is
        resolved by qualified-name lookup in the same module.
  \item \textbf{Class instantiation}: \texttt{ClassName(...)} is
        resolved to \texttt{ClassName.\_\_init\_\_}.
  \item \textbf{Method calls}: \texttt{self.m(...)} or \texttt{cls.m(...)}
        inside class bodies is resolved within the enclosing class scope.
\end{itemize}
When a call cannot be resolved exactly we fall back to short-name
matching against the qualified-name index, which recovers most
cross-module references while admitting a controlled false-positive rate.
Sibling edges $\mathcal{E}_{\mathrm{sib}}$ are constructed by
enumerating all pairs of methods belonging to the same class.

\subsection{Complexity Score: Caps and Weights}
\label{app:algo:hyperparams}

The complexity score $\hat{H}(v)$ defined in Eq.~\eqref{eq:complexity}
uses caps $(c_{\ell}, c_{c}, c_{d}) = (50, 10, 5)$ and weights
$(w_{\ell}, w_{c}, w_{d}) = (0.4, 0.4, 0.2)$. The cap $c_{\ell}=50$
on lines of code matches the median target length in our corpus; the
cap $c_c=10$ on McCabe complexity is roughly twice the median over our
filtered functions; the cap $c_d=5$ on nesting depth covers the tail of
practical Python code. The cap value of $2$ in $\phi(x,c)=\min(x/c,2)$
allows genuinely complex functions to score above the median without
letting outliers dominate.

\subsection{Inferability Score: Component Formulas}
\label{app:algo:inferability}

The five components of $\hat{I}(v)$ (Eq.~\ref{eq:inferability}) are
defined below. Default mixing weights are
$(\alpha,\beta,\gamma,\delta,\varepsilon) = (0.30, 0.25, 0.20, 0.10, 0.15)$.
All numerical constants used inside the components are listed in
Table~\ref{tab:inf-constants}; the formulas reference them directly.

\textbf{Caller signal $C_{\mathrm{caller}}$.}
For each in-file caller $u$ of $v$ we compute a per-caller specificity
$\mathrm{sp}(u,v)$ by scanning every call site inside $u$ that targets
$v$ (matched by short name). At each matching site we accumulate
$b_{\mathrm{site}}=0.50$ for the call itself, plus
$\eta_{\mathrm{lit}}=0.15$ per literal-constant argument,
$\eta_{\mathrm{name}}=0.05$ per name argument,
$\eta_{\mathrm{other}}=0.08$ per other expression argument,
$\eta_{\mathrm{kw}}=0.12$ per keyword argument, and a
$b_{\mathrm{found}}=0.10$ ``found'' bonus. The per-caller specificity is
clipped at $\mathrm{cap}_{\mathrm{caller}}=1.5$, with a fallback value
of $0.5$ when the caller does not actually invoke $v$ by name. Summing
over callers and normalizing,
$C_{\mathrm{caller}} = \min\!\left(\sum_u \mathrm{sp}(u,v) \,/\, n_{\mathrm{caller}},\, 1\right)$,
with normalization constant $n_{\mathrm{caller}}=3$.

\textbf{Callee signal $C_{\mathrm{callee}}$.}
$C_{\mathrm{callee}} = \min(|\mathrm{IntCallees}(v)|/n_{\mathrm{callee}},\,1)$
where $\mathrm{IntCallees}(v)$ is the set of intra-file functions that
$v$ calls, and $n_{\mathrm{callee}}=4$. A function that orchestrates
known helpers is more recoverable than one calling only opaque externals.

\textbf{Signature signal $C_{\mathrm{sig}}$.}
A monotone aggregate of four sub-bonuses, capped at $1$:
$+0.30$ if a return-type annotation is present, $+0.25$ if any parameter
carries a type annotation, $+\min(\mathrm{nameparts}/5,\,0.25)$ for the
descriptiveness of the function name (split on \texttt{\_}), and
$+\min(|\mathrm{params}_{\neg\mathrm{self}}|/6,\,0.20)$ for the count of
non-\texttt{self}/\texttt{cls} parameters.

\textbf{Documentation signal $C_{\mathrm{doc}}$.}
$C_{\mathrm{doc}}=0.5$ if a docstring is present, otherwise $0$. We do
not score docstring informativeness, both for robustness and because
Section~\ref{sec:method:cot} generates a richer textual rationale.

\textbf{Class signal $C_{\mathrm{class}}$.}
For methods, $C_{\mathrm{class}}=\min(\mathrm{sib}(v)/n_{\mathrm{sib}},\,1)$
where $n_{\mathrm{sib}}=5$, plus a $b_{\mathrm{init}}=0.30$ bonus when a
distinct \texttt{\_\_init\_\_} exists in the same class; the total is
clipped at $1$. Module-level functions receive $C_{\mathrm{class}}=0$.

\begin{table}[h]
\centering\small
\caption{Numerical constants inside the $\hat{I}$ components.}
\label{tab:inf-constants}
\begin{tabular}{llr}
\toprule
Component & Constant & Value \\
\midrule
\multirow{8}{*}{$C_{\mathrm{caller}}$}
  & call-site base $b_{\mathrm{site}}$              & $0.50$ \\
  & literal-arg increment $\eta_{\mathrm{lit}}$     & $0.15$ \\
  & name-arg increment $\eta_{\mathrm{name}}$       & $0.05$ \\
  & other-arg increment $\eta_{\mathrm{other}}$     & $0.08$ \\
  & keyword-arg increment $\eta_{\mathrm{kw}}$      & $0.12$ \\
  & found bonus $b_{\mathrm{found}}$                & $0.10$ \\
  & per-caller cap $\mathrm{cap}_{\mathrm{caller}}$ & $1.5$ \\
  & expected-callers norm $n_{\mathrm{caller}}$     & $3$ \\
\midrule
$C_{\mathrm{callee}}$ & expected-fan-out norm $n_{\mathrm{callee}}$ & $4$ \\
\midrule
\multirow{4}{*}{$C_{\mathrm{sig}}$}
  & return-type bonus       & $+0.30$ \\
  & param-type bonus        & $+0.25$ \\
  & name-parts term         & $\min(\mathrm{parts}/5,\,0.25)$ \\
  & non-self params term    & $\min(|p_{\neg\mathrm{self}}|/6,\,0.20)$ \\
\midrule
$C_{\mathrm{doc}}$    & docstring present & $0.5$ \\
\midrule
\multirow{2}{*}{$C_{\mathrm{class}}$}
  & sibling norm $n_{\mathrm{sib}}$           & $5$ \\
  & \texttt{\_\_init\_\_} bonus $b_{\mathrm{init}}$ & $0.30$ \\
\bottomrule
\end{tabular}
\end{table}

\subsection{Difficulty $\Delta$, Penalty $\rho$, and Hard Filters}
\label{app:algo:penalty}

The penalty $\rho(\Delta(v))$ in Eq.~\eqref{eq:fim_single} is built from
the residual difficulty $\Delta(v)$, defined as the share of complexity
that is not explained by context:
\begin{equation}
  \Delta(v) \;=\; \frac{\max\!\left(0,\,\hat{H}(v) - \hat{I}(v)\right)}{\hat{H}(v) + \epsilon}
  \;\in\; [0, 1).
  \label{eq:delta}
\end{equation}
The penalty $\rho$ is one-sided: targets with $\Delta(v) \le \tau_d$ are
left intact, while targets above the threshold are damped by a Gaussian
factor:
\begin{equation}
  \rho(\Delta) \;=\;
  \begin{cases}
    1, & \Delta \le \tau_d, \\[2pt]
    \exp\!\left(-\dfrac{(\Delta - \tau_d)^2}{2\sigma^2}\right), & \Delta > \tau_d.
  \end{cases}
  \label{eq:penalty}
\end{equation}
For single-function selection we use $\tau_d = 0.50$ and $\sigma = 0.20$.
The asymmetry reflects an asymmetry in the underlying objective:
trivially easy targets are already eliminated by the hard filters below,
so we do not need to additionally reward high inferability; conversely,
targets that remain hard even with full context become unlearnable noise
and must be down-weighted.

\textbf{Hard filters.}
Independent of the smooth score, we discard \emph{(i)} files outside
the line-count window $[L_{\min}, L_{\max}] = [50, 1800]$ entirely;
\emph{(ii)} functions with $\mathrm{LoC}(v) \notin [10, 200]$ to control
training-instance length; \emph{(iii)} dunder methods
(\texttt{\_\_init\_\_}, \texttt{\_\_repr\_\_}, etc.), which serve as
scaffolding rather than reasoning targets; \emph{(iv)} functions with
$\hat{H}(v) < \tau_{\hat H} = 0.15$ to remove trivial cases; and
\emph{(v)} functions with $\mathrm{FIM}(v) < \tau_{\mathrm{FIM}} = 0.08$.
A file with no surviving target contributes nothing to the corpus.

\subsection{Worked Example: Scoring \texttt{Calculator.total}}
\label{app:algo:worked_example}

To make the numbers in Figure~\ref{fig:pdg_score}(b) reproducible, this
subsection walks through every quantity that goes into $\hat{H}$,
$\hat{I}$, and the final FIM score for the running example. The target
function is:

\begin{center}\small
\begin{minipage}{0.78\linewidth}
\begin{verbatim}
def total(self) -> int:
    """Sum positive integer entries."""
    s, n = 0, 0
    for v in self.history:
        if is_int(v):
            if v > 0:
                s = add(s, v)
                n += 1
    if n == 0: return 0
    return s
\end{verbatim}
\end{minipage}
\end{center}

\textbf{Lines of code (LoC = 10).}
Following Python's \texttt{ast} convention,
$\mathrm{LoC}(v)=\texttt{end\_lineno}-(\texttt{lineno}-1)$, the number
of source lines from the \texttt{def} keyword through the last
statement, both inclusive. Counting the ten lines above (including the
docstring) gives $\mathrm{LoC}=10$.

\textbf{Cyclomatic complexity (CC = 5).}
The McCabe counter starts at $1$ and adds $1$ per branching node:
\texttt{If}/\texttt{IfExp}, \texttt{For}/\texttt{While},
\texttt{ExceptHandler}, $(\text{values}{-}1)$ per \texttt{BoolOp}, and
one per generator inside a comprehension. For
\texttt{Calculator.total}: base ($+1$), the \texttt{for} loop ($+1$),
\texttt{if is\_int} ($+1$), \texttt{if v>0} ($+1$), \texttt{if n==0}
($+1$), totalling $\mathrm{CC}=5$. Tuple unpacking and \texttt{n+=1}
are not branches.

\textbf{Maximum nesting depth (D = 3).}
Depth increments only on structural containers (\texttt{If},
\texttt{For}, \texttt{While}, \texttt{AsyncFor}, \texttt{With},
\texttt{AsyncWith}, \texttt{Try}). The deepest path in
\texttt{total} is \texttt{for}~$\to$~\texttt{if is\_int}~$\to$~%
\texttt{if v>0}, giving $\mathrm{D}=3$.

\textbf{Complexity score $\hat{H}=0.40$.}
Plugging into Eq.~\ref{eq:complexity} with the default
$(w_{\ell},w_{c},w_{d})=(0.4,0.4,0.2)$ and
$(c_{\ell},c_{c},c_{d})=(50,10,5)$:
\begin{align*}
w_{\ell}\,\phi(10,50) &= 0.4\!\cdot\!0.20 = 0.08, \qquad
w_{c}\,\phi(5,10)     = 0.4\!\cdot\!0.50 = 0.20, \\
w_{d}\,\phi(3,5)      &= 0.2\!\cdot\!0.60 = 0.12, \quad
\hat{H}(\texttt{total}) = 0.08+0.20+0.12 = 0.40.
\end{align*}

\textbf{Inferability score $\hat{I}=0.48$.}
\emph{(i)~Caller specificity.} The only in-file caller is
\texttt{Calculator.mean}, which invokes \texttt{self.total()} with no
arguments and no keywords. Per
Table~\ref{tab:inf-constants},
$\mathrm{sp}=b_{\mathrm{site}}+b_{\mathrm{found}}=0.50+0.10=0.60$.
Aggregating and normalizing by $n_{\mathrm{caller}}=3$,
$C_{\mathrm{caller}}=\min(0.60/3,1.0)=0.20$ and
$\alpha\,C_{\mathrm{caller}}=0.06$.
\emph{(ii)~Callee count.} \texttt{total} calls \texttt{is\_int} and
\texttt{add} in-file (\texttt{self.history.append(v)} does not count
because \texttt{append} is not in $\mathcal{V}$), so
$C_{\mathrm{callee}}=\min(2/n_{\mathrm{callee}},1.0)=\min(2/4,1.0)=0.50$
and $\beta\,C_{\mathrm{callee}}=0.13$.
\emph{(iii)~Signature.} Sub-bonuses $+0.30$ for \texttt{-> int};
$+\min(1/5,0.25)=0.20$ for the descriptive name \texttt{total};
$0$ for parameter type annotations; $0$ for non-\texttt{self} parameters:
$C_{\mathrm{sig}}=0.50$ and $\gamma\,C_{\mathrm{sig}}=0.10$.
\emph{(iv)~Documentation.} A docstring is present, so
$C_{\mathrm{doc}}=0.5$ and $\delta\,C_{\mathrm{doc}}=0.05$.
\emph{(v)~Class context.} \texttt{total} sits among three siblings; the
base score is $\min(3/n_{\mathrm{sib}},1.0)=\min(3/5,1.0)=0.60$, plus the
$b_{\mathrm{init}}=0.30$ bonus for the distinct \texttt{\_\_init\_\_} of
\texttt{Calculator}, giving $C_{\mathrm{class}}=0.90$ and
$\varepsilon\,C_{\mathrm{class}}=0.14$.
Summing: $\hat{I}(\texttt{total})=0.06+0.13+0.10+0.05+0.14=0.48$.

\textbf{Final FIM score and selection.}
$\Delta(\texttt{total})=\max(0,-0.08)/0.40=0$, so
$\rho(\Delta)=1$ and
$\mathrm{FIM}(\texttt{total})=(0.40\!\cdot\!0.48)/(0.40\!+\!0.48)\approx 0.22 \geq \tau_{\mathrm{FIM}}=0.08$,
selecting it as a mask target. The four short helpers are removed before
scoring by the LoC hard filter; \texttt{Calculator.\_\_init\_\_} is
removed by the dunder filter.

\subsection{Multi-Function Group Score}
\label{app:algo:groups}

\textbf{Group score.}
For a group $G = \{v_1, \dots, v_k\}$ enumerated from one of eight
topology patterns,
\begin{equation}
  \mathrm{FIM}(G) \;=\;
  \mathrm{Coup}(G)
  \cdot
  \frac{\hat{H}(G)\,\hat{I}(G)}{\hat{H}(G) + \hat{I}(G) + \epsilon}
  \cdot
  \rho^G\!\left(\Delta(G)\right),
  \label{eq:fim_group}
\end{equation}
where $\hat{H}(G)=\tfrac{1}{k}\sum_v \hat{H}(v)$ averages individual
complexities, $\Delta(G)$ is defined analogously to the single-function
case, $\rho^G$ uses group-specific parameters $\tau_d^G=0.55$ and
$\sigma^G=0.20$ (slightly more permissive than the single-function
$\tau_d=0.50$, reflecting that joint masking inherently lowers $\hat{I}$),
and the coupling term $\mathrm{Coup}(G)\in[0,1]$ is a normalized count
of intra-group edges plus a shared-state ratio:
\begin{equation}
  \mathrm{Coup}(G)
  \;=\; w_{\mathrm{c}}\,\frac{|E^{\text{call}}_G|}{k(k\!-\!1)}
   + w_{\mathrm{s}}\,\frac{|E^{\text{sib}}_G|}{\binom{k}{2}}
   + w_{\mathrm{st}}\,\overline{\mathrm{Jacc}}_G,
  \label{eq:coupling}
\end{equation}
with $\overline{\mathrm{Jacc}}_G$ the mean Jaccard similarity over pairs
in $G$ between sets of accessed instance attributes (\texttt{self.x}).
We use $(w_{\mathrm{c}}, w_{\mathrm{s}}, w_{\mathrm{st}}) = (0.50,
0.20, 0.30)$. The coupling term enters multiplicatively because, unlike
the single-function case, the value of a multi-function target derives
specifically from cross-function dependencies; a weakly coupled group
is indistinguishable from $k$ independent samples.

\textbf{Group inferability.}
$\hat{I}(G)$ is recomputed under the assumption that \emph{all} members
of $G$ are masked simultaneously: caller, callee, and sibling
contributions from inside $G$ are excluded; docstrings of group members
contribute zero (they are part of the masked body); and the
\texttt{\_\_init\_\_} bonus in $C_{\mathrm{class}}$ is awarded only when
\texttt{\_\_init\_\_} is itself outside $G$. Signature information is
retained because we do not mask function signatures. The remaining
constants $(b_{\mathrm{site}}, \eta_{\mathrm{lit}}, \dots)$ in
Table~\ref{tab:inf-constants} are unchanged. This re-computation
prevents groups from receiving spurious credit for intra-group
references that disappear under joint masking.

\textbf{Topology taxonomy.}
A candidate group must form a connected subgraph in
$\mathcal{E}_{\mathrm{call}} \cup \mathcal{E}_{\mathrm{sib}}$. We
enumerate eight patterns:
\begin{itemize}\setlength\itemsep{0.1em}
  \item \emph{$k=2$}: \emph{caller-callee} (direct $A\!\to\!B$ edge);
  \emph{co-callee} (two callees of a common in-file caller);
  \emph{sibling-coupled} (same-class methods sharing $\geq 1$ instance
  attribute); \emph{mutual-call} ($A\!\rightleftharpoons\!B$).
  \item \emph{$k=3$}: \emph{call-chain} ($A\!\to\!B\!\to\!C$); \emph{hub}
  ($A\!\to\!B,\ A\!\to\!C$); \emph{fan-in} ($B\!\to\!A,\ C\!\to\!A$);
  \emph{class-triad} (three pairwise state-sharing methods of the same
  class).
\end{itemize}

\textbf{Group selection.}
We score every candidate group emitted by the topology enumerator and
discard those failing any of the following defaults:
each member must satisfy $\mathrm{LoC} \in [10,200]$ and not be a dunder;
$\mathrm{Coup}(G) \ge \tau^G_{\mathrm{coup}}=0.15$;
$\hat{H}(G) \ge \tau^G_{\hat H}=0.15$; total LoC ratio of the group
relative to the file does not exceed
$\theta_2=0.30$ for $k=2$ or $\theta_3=0.40$ for $k=3$; and the score
clears a size-specific floor,
$\mathrm{FIM}(G) \ge \tau^G_{\mathrm{FIM,2}}=0.04$ for pairs and
$\mathrm{FIM}(G) \ge \tau^G_{\mathrm{FIM,3}}=0.03$ for triples.
Remaining candidates are sorted by $\mathrm{FIM}(G)$ and selected
greedily under non-overlap: a function may belong to at most one selected
group per file. Per-file caps $(N_2, N_3) = (5, 3)$ prevent any single
file from dominating the corpus.

\begin{algorithm}[h]
\caption{Multi-function group FIM target selection for one file.}
\label{alg:selection_multi}
\begin{algorithmic}[1]
  \Require Source file $s$; per-function scores $\hat{H}(v),\hat{I}(v)$;
           thresholds $\tau^G_{\mathrm{coup}}=0.15,\;\tau^G_{\hat H}=0.15,\;
           \tau^G_{\mathrm{FIM,2}}=0.04,\;\tau^G_{\mathrm{FIM,3}}=0.03$;
           LoC ratio caps $(\theta_2,\theta_3)=(0.30,0.40)$;
           per-file caps $(N_2, N_3)=(5, 3)$
  \State $\mathcal{V}, \mathcal{E}_{\mathrm{call}}, \mathcal{E}_{\mathrm{sib}} \gets \mathrm{BuildPDG}(\mathrm{ParseAST}(s))$
  \State $\mathcal{G} \gets \emptyset$
  \For{each topology pattern $P \in \{\text{caller-callee, co-callee, sibling-coupled, mutual-call,}$
       $\quad\text{call-chain, hub, fan-in, class-triad}\}$}
    \State Enumerate all connected subgraphs $G \subseteq \mathcal{V}$ matching $P$
           in $\mathcal{E}_{\mathrm{call}}\cup\mathcal{E}_{\mathrm{sib}}$
    \For{each candidate $G$}
      \If{any $v\!\in\!G$ fails the per-function filter
          \textbf{or} $\mathrm{LoC\_ratio}(G,s)>\theta_{|G|}$
          \textbf{or} $\mathrm{Coup}(G)<\tau^G_{\mathrm{coup}}$
          \textbf{or} $\hat{H}(G)<\tau^G_{\hat H}$}
        \State \textbf{continue}
      \EndIf
      \State Recompute $\hat{I}(G)$ under joint masking
            \Comment{exclude intra-group references; $C_{\mathrm{doc}}\!=\!0$}
      \State $\Delta(G) \gets \max(0, \hat{H}(G)-\hat{I}(G))/(\hat{H}(G)+\epsilon)$
      \State $\mathrm{FIM}(G) \gets \mathrm{Coup}(G)\cdot
              \dfrac{\hat{H}(G)\,\hat{I}(G)}{\hat{H}(G)+\hat{I}(G)+\epsilon}\cdot \rho^G(\Delta(G))$
      \If{$\mathrm{FIM}(G) \ge \tau^G_{\mathrm{FIM},|G|}$}
        \State $\mathcal{G} \gets \mathcal{G} \cup \{(G,P,\mathrm{FIM}(G))\}$
      \EndIf
    \EndFor
  \EndFor
  \State Sort $\mathcal{G}$ by $\mathrm{FIM}(G)$ descending
  \State $\mathcal{S} \gets \emptyset$
  \For{each $(G,P,\mathrm{FIM}(G)) \in \mathcal{G}$}
    \If{any $v \in G$ already covered by $\mathcal{S}$
        \textbf{or} per-size cap $(N_2$ for $k\!=\!2$, $N_3$ for $k\!=\!3)$ exceeded}
      \State \textbf{continue}
    \EndIf
    \State $\mathcal{S} \gets \mathcal{S} \cup \{G\}$
  \EndFor
  \State \Return $\mathcal{S}$
\end{algorithmic}
\end{algorithm}

\subsection{Worked Example: A \texttt{caller-callee} Pair}
\label{app:algo:groups_example}

To illustrate the group score, consider a small file with two
top-level functions:

\begin{center}\small
\begin{minipage}{0.82\linewidth}
\begin{verbatim}
def normalize(text: str) -> str:
    """Lowercase and strip punctuation."""
    out = []
    for ch in text:
        if ch.isalnum() or ch.isspace():
            out.append(ch.lower())
    return "".join(out)

def word_freq(doc: str) -> dict[str, int]:
    """Count tokens of `doc` after normalization."""
    cleaned = normalize(doc)
    counts = {}
    for tok in cleaned.split():
        counts[tok] = counts.get(tok, 0) + 1
    return counts
\end{verbatim}
\end{minipage}
\end{center}

\textbf{Topology and pre-checks.} The pair
$G\!=\!\{\texttt{normalize}, \texttt{word\_freq}\}$ matches the
\emph{caller-callee} pattern with one call edge
$\texttt{word\_freq}\!\to\!\texttt{normalize}$. They are top-level
(no class), so the sibling component of $\mathrm{Coup}$ is zero and
the shared-state component is zero (no \texttt{self.\_}~accesses). The
combined LoC is roughly $14$ lines, well under the $\theta_2\!=\!0.30$
file-ratio cap.

\textbf{Coupling.} With one call edge in a pair,
$|E^{\text{call}}_G|/(k(k\!-\!1))=1/2$. Plugging into
Eq.~\ref{eq:coupling} with $(w_c,w_s,w_{st})=(0.50,0.20,0.30)$:
$\mathrm{Coup}(G)=0.50\cdot 0.5+0+0=0.25$, comfortably above the
$\tau^G_{\mathrm{coup}}=0.15$ floor.

\textbf{Group $\hat{H}$.} Per-function complexities (computed as in
Section~\ref{sec:method:complexity}) come out to roughly
$\hat{H}(\texttt{normalize})\!\approx\!0.36$ and
$\hat{H}(\texttt{word\_freq})\!\approx\!0.30$,
giving $\hat{H}(G)\!\approx\!0.33$.

\textbf{Group $\hat{I}$ under joint masking.} Both functions are
masked, so the call edge between them no longer contributes to
$C_{\mathrm{caller}}$ or $C_{\mathrm{callee}}$. Signature information
is retained: both have descriptive names and full type annotations
(strong $C_{\mathrm{sig}}$); docstrings vanish under joint masking
($C_{\mathrm{doc}}=0$). With no surviving caller/callee/class signal,
the group inferability collapses to $\hat{I}(G)\!\approx\!0.18$,
dominated by the signature term.

\textbf{Final group FIM.} The harmonic-mean-like product
$\hat{H}\hat{I}/(\hat{H}+\hat{I})\!\approx\!(0.33\!\cdot\!0.18)/0.51\!\approx\!0.116$;
$\Delta(G)\!=\!(0.33-0.18)/0.33\!\approx\!0.45<\tau_d^G=0.55$ so
$\rho^G=1$; and $\mathrm{Coup}(G)=0.25$:
$\mathrm{FIM}(G)\approx 0.25\!\times\!0.116\!\approx\!0.029$. This
example sits below the pair floor $\tau^G_{\mathrm{FIM,2}}\!=\!0.04$ and
would therefore \emph{not} be selected---the toy module is too small
for the joint-masking inferability to remain large enough. Real corpus
pairs that pass the threshold typically come from class-method pairs
with non-trivial sibling and shared-state coupling, where $\hat{I}(G)$
retains substantial $C_{\mathrm{class}}$ contribution even under joint
masking. The walkthrough nonetheless shows the mechanism: joint masking
strictly reduces $\hat{I}$ relative to the single-function average,
and the coupling factor must compensate.

\subsection{Chain-of-Thought Generation Prompt Template}
\label{app:algo:cot-gen}
Each FIM rationale--implementation pair is produced by Gemini-3-Flash
in a single forward pass: given a Python source file with one function
body redacted, the model is asked to first reason about what the body
should do and then write a candidate implementation. Crucially, the
model is \emph{not} shown the ground-truth body at this stage; both
the rationale and the predicted body must be derived from the
surrounding context alone. The (rationale, predicted body) pairs that
survive the quality filter (Appendix~\ref{app:algo:filter}) are then
used to assemble mid-training samples
(Appendix~\ref{app:algo:cot-format}).
\begin{quote}\small
\textit{[System]} You are an expert Python programmer.

\noindent\textit{[User]} Below is a Python file where one function's
body has been replaced with \texttt{\# <MASKED\_FUNCTION\_BODY>}.
Complete the masked function from the surrounding context.

\noindent\textit{Procedure.} Analyze the visible context---imports,
sibling functions, classes, call sites, and helpers invoked by the
target---then reason step by step about what the function must do
given its signature, type hints, and usage pattern. Only after this
analysis, write an implementation consistent with the file's coding
style.

\noindent\textit{Output format.} Return two sections in order:
\texttt{\#\#\# Reasoning} (a step-by-step natural-language analysis)
and \texttt{\#\#\# Implementation} (a fenced \texttt{python} block
containing the function body only, without the signature).

\noindent\texttt{<file with target body redacted>}

\noindent The function to complete: \texttt{<function name>}.
\end{quote}
For brevity, the skeleton condenses three pieces of the production
prompt that govern output formatting rather than the task itself:
(i) the explicit three-step task breakdown (analyze context, reason,
implement) and the four sub-bullets under the reasoning step
(signature semantics, intra-file usage, callee dependencies, overall
purpose), which we collapse into a single \textit{Procedure} paragraph
and which direct the model toward the same context sources scored by
$\hat{I}$ (Section~\ref{sec:method:inferability}); (ii) the
formatting checklist enforcing body-only output, four-space
indentation, and explicit docstring handling; and (iii) the verbatim
Markdown headers and code-fence delimiters that make the two sections
programmatically separable. The condensed material specifies
\emph{how} the response is formatted but does not change \emph{what}
is being elicited. We sample with the Gemini API's near-greedy
configuration (\texttt{temperature}~$=0$, \texttt{topK}~$=1$); the
verbatim prompt and all post-processing logic will be released with
the corpus to support exact replication.

\subsection{Completion-Quality Filtering Prompt Template}
\label{app:algo:filter}
Each candidate FIM sample, paired with a trial completion, is screened
by an LLM judge that assesses both whether the masked function is
recoverable from its surrounding context and the quality of the
completion against the ground truth. Filtering is performed by
Gemini-3-Flash using the following prompt skeleton:
\begin{quote}\small
\textit{[System]} You are an expert code reviewer.

\noindent\textit{[User]} A model was asked to complete a masked
function from its surrounding code context. Evaluate the result along
two axes: (i) whether the function is feasible to complete from
context alone, and (ii) the quality of the completion compared with
the ground truth.

\noindent The source file with the target body redacted (shown as
\texttt{<MASKED>}):

\noindent\texttt{<file with target body redacted>}

\noindent The function name, the ground-truth body, and the model's
completion:

\noindent\texttt{<function name>}, \texttt{<ground-truth body>},
\texttt{<model completion>}

\noindent\textit{Part 1 -- Feasibility.} Mark the function
\emph{infeasible} if it depends on uncommon external APIs, external
conventions, or magic constants that cannot be inferred from the
visible context; otherwise mark it \emph{feasible}.

\noindent\textit{Part 2 -- Quality.} Score the completion on a
$1$--$5$ scale along five dimensions---\emph{correctness},
\emph{executability}, \emph{API usage}, \emph{readability}, and
\emph{completeness}---each with a one-sentence justification, then
assign an overall $1$--$5$ score.

\noindent\textit{Output.} Return a JSON object containing the
feasibility verdict, the five component scores with reasons, the
overall score, and a \texttt{should\_discard} flag set to
\texttt{true} when the function is infeasible, when
\texttt{executability} $= 1$, or when both \texttt{overall\_score}
$\le 2$ and \texttt{executability} $\le 2$.
\end{quote}
For brevity, the skeleton above omits three pieces of the production
prompt that anchor the scores rather than alter the procedure: (i)
the per-level rubric for each of the five dimensions (e.g.,
\emph{correctness} $5$ = ``functionally identical,'' $3$ = ``partially
correct, some important cases wrong,'' $1$ = ``completely
incorrect''), which fixes a consistent scale across samples; (ii) the
expanded feasibility checklist enumerating concrete infeasibility
factors (reliance on niche libraries, knowledge of external
conventions, insufficient context, system-specific magic constants)
and corresponding feasibility indicators; and (iii) the full JSON
schema with named fields for every component reason, key
similarities/differences, and a free-text \texttt{discard\_reason}.
The omitted material specifies \emph{how} each axis is graded but
does not change \emph{what} is graded; We sample with the Gemini API's near-greedy configuration
(\texttt{temperature}~$=0$, \texttt{topK}~$=1$). The verbatim prompt together with all post-processing logic
will be released with the corpus to support exact replication.

\subsection{Mid-Training Sample Formatting Template}
\label{app:algo:cot-format}
After upstream CoT generation (Appendix~\ref{app:algo:cot-gen}) and
quality filtering (Appendix~\ref{app:algo:filter}), each retained
sample is serialized into a chat-style turn pair: the user turn poses
the FIM completion task on the file with one function body redacted,
and the assistant turn carries Gemini's rationale followed by
Gemini's predicted body, both produced in the upstream pass. The
ground-truth body from the source repository is used only by the
filter to admit the sample; it does not appear in the training
target.
\begin{quote}\small
\textit{[System]} You are an expert Python programmer.

\noindent\textit{[User]} Below is a Python file where one function's
body has been replaced with \texttt{\# <MASKED\_FUNCTION\_BODY>}.
Complete the masked function from the surrounding context.
\noindent\textit{[Procedure and output-format instructions identical
to Appendix~\ref{app:algo:cot-gen}.]}

\noindent\texttt{<file with target body redacted>}

\noindent The function to complete: \texttt{<function name>}.

\noindent\textit{[Assistant]}\\
\texttt{\#\#\# Reasoning}\\
\texttt{<Gemini-produced rationale>}\\
\texttt{\#\#\# Implementation}\\
\texttt{```python}\\
\texttt{<Gemini-produced implementation>}\\
\texttt{```}
\end{quote}

\subsection{Negative-Observation Patterns}
\label{app:algo:neg_patterns}

The recovery-rate analysis (Section~\ref{sec:analysis:recovery}) flags
a tool output as a \emph{negative observation} when it matches any of
the following patterns: Python stack-trace prefixes
(\texttt{Traceback (most recent call last):}); error class names
(\texttt{SyntaxError}, \texttt{IndentationError}, \texttt{NameError},
\texttt{ImportError}, \texttt{AttributeError}, \texttt{TypeError},
\texttt{ValueError}); shell error markers
(\texttt{command not found}, \texttt{permission denied}, \texttt{No such file or directory});
and harness-specific failure strings
(\texttt{No replacement was performed}, \texttt{Patch did not apply},
\texttt{tests failed}). Patterns are matched case-insensitively.

\section{Training Hyperparameters}
\label{app:hp}

We use \textsc{LlamaFactory} for FIM mid-training, R2E-Gym
post-training, and SWE-Lego post-training, and \textsc{torchtune} for
SWE-Smith post-training. All runs use AdamW with bf16 mixed precision
and a cosine learning-rate schedule. For R2E-Gym and SWE-Smith we follow
the official release scripts; for SWE-Lego we use $2$ epochs instead of
the official $4$ to prevent overfitting (all other hyperparameters
unchanged). Tables~\ref{tab:hp-midtrain} and~\ref{tab:hp-posttrain} list
the exact settings; effective (global) batch sizes assume $8$ GPUs.

\begin{table}[h]
\centering
\footnotesize
\caption{FIM mid-training hyperparameters, applied uniformly to all
three base models (Qwen2.5-Coder-7B-Instruct,
Qwen2.5-Coder-14B-Instruct, and Qwen3-8B).}
\label{tab:hp-midtrain}
\setlength{\tabcolsep}{8pt}
\begin{tabular}{ll}
\toprule
\textbf{Hyperparameter} & \textbf{Value} \\
\midrule
Optimizer             & AdamW \\
Learning rate         & $1.0\!\times\!10^{-5}$ \\
LR schedule           & Cosine \\
Warmup ratio          & $0.1$ \\
Weight decay          & $0.05$ \\
Epochs                & $1$ \\
Per-device batch size & $1$ \\
Gradient accumulation & $16$ \\
Effective batch size  & $128$ \\
Sequence length       & $32{,}768$ \\
Precision             & bf16 \\
\bottomrule
\end{tabular}
\end{table}

\begin{table}[h]
\centering
\footnotesize
\caption{Agentic post-training hyperparameters for the three pipelines.
R2E-Gym and SWE-Smith follow their official released scripts; SWE-Lego
follows the official recipe except for the epoch count.}
\label{tab:hp-posttrain}
\setlength{\tabcolsep}{6pt}
\begin{tabular}{lccc}
\toprule
\textbf{Hyperparameter} & \textbf{R2E-Gym} & \textbf{SWE-Smith} & \textbf{SWE-Lego} \\
\midrule
Base model            & Qwen2.5-Coder-7B-Inst.
                      & Qwen2.5-Coder-7B-Inst.
                      & Qwen3-8B (FIM-midtrained) \\
Optimizer             & AdamW & AdamW (fused) & AdamW \\
Learning rate         & $1.0\!\times\!10^{-5}$
                      & $1.0\!\times\!10^{-4}$
                      & $1.0\!\times\!10^{-4}$ \\
LR schedule           & Cosine & Cosine & Cosine \\
Warmup                & ratio $0.05$ & $5$ steps & ratio $0.1$ \\
Weight decay          & $0.0$ & $0.01$ & $0.01$ \\
Epochs                & $2$ & $3$ & $2$\textsuperscript{$\dagger$} \\
Per-device batch size & $1$ & $1$ & $1$ \\
Gradient accumulation & $1$ & $4$ & $8$ \\
Effective batch size  & $8$ & $32$ & $64$ \\
Sequence length       & $32{,}768$ & $32{,}768$ & $131{,}072$ \\
Precision             & bf16 & bf16 & bf16 \\
\bottomrule
\end{tabular}
\\[0.3em]
{\raggedright \footnotesize \textsuperscript{$\dagger$}Deviates from the
official SWE-Lego recipe ($4$ epochs); reduced to $2$ to prevent
overfitting on the FIM-midtrained Qwen3-8B base.\par}
\end{table}

\subsection{Compute Resources}
\label{app:hp:compute}
All experiments are run on a single node of $8$ NVIDIA H100 80\,GB GPUs.
Reproducing the full set of results in this paper---FIM mid-training
of three base models (Qwen2.5-Coder-7B/14B-Instruct and Qwen3-8B), the
three agentic post-training pipelines (R2E-Gym, SWE-Smith, SWE-Lego)
and their corresponding mid-trained variants, the data/CoT/granularity
ablations of Section~\ref{sec:exp:ablation}, and the multi-seed evaluation
sweeps---takes roughly $30$ days of wall-clock time on the $8$-GPU node,
i.e.\ $\approx\!5{,}760$ GPU-hours in total. Preliminary and discarded
runs not reported in this paper account for an additional ${\sim}30\%$
of compute.

\section{Extended Behavioral Analysis (SWE-Bench-Verified)}
\label{app:behavior_extra}

This appendix collects the trajectory-level material that supplements
the headline analysis in Section~\ref{sec:analysis}. Unless stated
otherwise, statistics are run-means over three independent evaluation
runs of each checkpoint on SWE-Bench-Verified ($500$ instances per run).

\subsection{Pass Rate by Gold-Patch Shape}
\label{app:behavior_extra:patchshape}

\begin{figure}[h]
\centering
\includegraphics[width=0.62\linewidth]{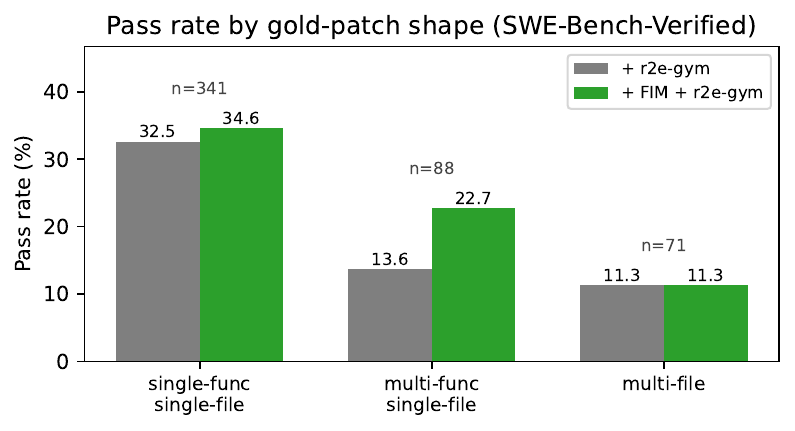}
\caption{Pass rate on SWE-Bench-Verified ($14$B + R2E-Gym) stratified
by gold-patch shape, run-means over three evaluation runs per
checkpoint.}
\label{fig:patchshape}
\end{figure}

Figure~\ref{fig:patchshape} reports the pass-rate stratification used
in Section~\ref{sec:analysis:multifn}: the largest gain
($+9.1$~pp) is on multi-function single-file tasks
($n{=}88$), more than $4{\times}$ the $+2.1$~pp gain on single-function
tasks ($n{=}341$); multi-file tasks ($n{=}71$) remain hard for both
checkpoints ($\sim\!11.3\%$ each).

\subsection{Full Trajectory Metrics}
\label{app:behavior_extra:metrics}

Table~\ref{tab:behavior_full} reports the full trajectory metrics
summarized in Table~\ref{tab:behavior_main}, including unsolved-task
breakdowns and output/context summaries.

\begin{table}[h]
\centering
\caption{Trajectory-level behavioral metrics on SWE-Bench-Verified
(14B, R2E-Gym), run-means over three evaluation runs of each
checkpoint. Steps, edits, and errors-encountered are means per task.
``Empty patch'' is the fraction of \emph{failed} trajectories whose
final \texttt{output\_patch} is empty; ``Loc.\ correct'' is the
fraction of all trajectories that locate at least one file overlapping
the gold patch; ``Ctx @ success'' is the mean prompt context length on
solved trajectories.}
\label{tab:behavior_full}
\small
\setlength{\tabcolsep}{4pt}
\begin{tabular}{l c c c c c c c}
\toprule
& \multicolumn{2}{c}{Steps / task}
& \multicolumn{2}{c}{Edits / task}
& \multicolumn{2}{c}{Errors enc.}
& Recovery \\
\cmidrule(lr){2-3}\cmidrule(lr){4-5}\cmidrule(lr){6-7}
Setting & solved & unsolved & solved & unsolved & solved & unsolved & rate (\%) \\
\midrule
+ R2E-Gym
  & 15.1 & 22.4 & 3.3 & 5.4 & 3.5 & 5.8 & 24.8 \\
\textbf{+ FIM-Midtrain + R2E-Gym}
  & \textbf{23.6} & \textbf{32.7} & \textbf{7.4} & \textbf{10.9}
  & \textbf{6.2} & \textbf{8.1} & \textbf{28.8} \\
\midrule
\multicolumn{8}{l}{\textit{Output and context summaries}} \\
\midrule
& \multicolumn{2}{c}{Empty patch (\% failed)}
& \multicolumn{2}{c}{Loc.\ correct (\% all)}
& \multicolumn{2}{c}{Ctx @ success (tok)}
& Pass (\%) \\
\cmidrule(lr){2-3}\cmidrule(lr){4-5}\cmidrule(lr){6-7}
+ R2E-Gym
  & \multicolumn{2}{c}{$3.0$} & \multicolumn{2}{c}{$70.6$}
  & \multicolumn{2}{c}{$11{,}826$} & 26.2 \\
\textbf{+ FIM-Midtrain + R2E-Gym}
  & \multicolumn{2}{c}{$\mathbf{0.3}$} & \multicolumn{2}{c}{$\mathbf{73.4}$}
  & \multicolumn{2}{c}{$\mathbf{16{,}397}$} & \textbf{29.2} \\
\bottomrule
\end{tabular}
\end{table}

\subsection{Iterate-and-Verify: Action-Type Distribution}
\label{app:behavior_extra:actions}

Mid-training shifts the action-type distribution toward an
iterate-and-verify policy. The share of \texttt{search} actions falls
from $15.3\%$ to $11.0\%$, while the share of \texttt{execute\_bash}
actions rises from $19.5\%$ to $24.6\%$: the agent spends a smaller
fraction of its budget on passive repository lookup and a larger
fraction on actively running scripts and tests, then refining its
edit. The cost is more steps---ours hits the step cap on $\sim\!45\%$
of trajectories vs.\ $7\%$ for the baseline---but this cost is paid on
the unsolved tail; on the solved set the additional steps translate
into more correct patches.

\subsection{Failure-Mode Breakdown}
\label{app:behavior_extra:failure}

\begin{figure}[h]
\centering
\includegraphics[width=0.55\linewidth]{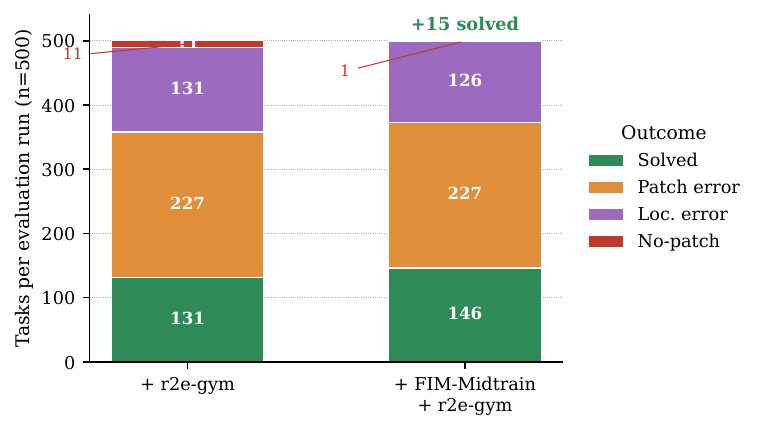}
\caption{Outcome distribution per evaluation run on SWE-Bench-Verified
(14B, R2E-Gym), averaged over three runs.}
\label{fig:failure_app}
\end{figure}

We label every failed trajectory using a deterministic patch-quality
cascade: \emph{no-patch} (output diff is empty); \emph{localization
error} (non-empty patch but zero file overlap with the gold patch);
\emph{patch error} (file overlap with gold patch but tests fail).
Figure~\ref{fig:failure_app} reports the outcome distribution: across
the three runs, mid-training cuts no-patch failures by an order of
magnitude ($\sim\!11\!\to\!\sim\!1$ trajectories per run), trims
localization errors slightly ($\sim\!131\!\to\!\sim\!126$), and leaves
the patch-error count essentially flat ($\sim\!227\!\to\!\sim\!227$).
The $+15$-task gain in solved count is therefore primarily driven by
the no-patch mode collapse.

\subsection{No-Patch Mode: Mechanism}
\label{app:behavior_extra:nopatch}

The baseline no-patch failures recovered by mid-training are
distributed across all task buckets and are not concentrated in any
single repository. In every recovered case the baseline emitted
\texttt{<finish>} without applying any \texttt{str\_replace}, while
ours applied at least one edit. We interpret this as a direct
consequence of the FIM training signal: under FIM the model is always
conditioned to produce a non-empty token span between the prefix and
suffix, and this disposition survives the post-training pipeline.

\subsection{Multi-File Tasks Are Not Differentially Helped}
\label{app:behavior_extra:multifile}

On the $71$ Verified tasks whose gold patch spans $\geq 2$ files, both
checkpoints solve $\sim\!11.3\%$ on average and per-instance
head-to-head on this bucket is balanced. We attribute this to a
granularity mismatch: our function-aware FIM mid-training operates at
the function level within files, so cross-file coordination is not
directly trained for. Extending the selection algorithm to cross-file
function pairs is a natural follow-up.

\subsection{A Concrete Contrast}
\label{app:behavior_extra:contrast}

The shift in termination behavior is clearest on cases where the
baseline quits prematurely. On \texttt{scikit-learn-26323}, for
instance, the baseline trajectory runs for a single step in every run,
immediately emits \texttt{<finish>} with an empty edit and the wrong
target file, and is scored as a failure; on the same task our agent
runs $\sim\!41$ steps, applies on the order of $20$
\texttt{str\_replace} operations, observes multiple negative tool
returns, and recovers to a passing patch. Pooling across the three
runs, on the order of $\sim\!16$ of the $\sim\!55$ tasks that ours
uniquely solves on Verified have this signature: the baseline emits
\texttt{<finish>} with zero file overlap with the gold patch (often
before any negative observation has even arrived), while ours iterates
past intermediate signals and converges on a correct edit.
Mid-training does not enable a categorically new capability on these
instances; it changes the agent's stopping policy.

\section{Behavioral Analysis on SWE-Bench-Lite}
\label{app:behavior_lite}

This appendix replicates the Verified analysis on the SWE-Bench-Lite
test split ($300$ instances). The same trajectory-parsing pipeline and
failure-mode taxonomy are used; statistics are run-means over three
independent evaluation runs of each checkpoint.

The qualitative trends from Verified hold on Lite: mid-training raises
the recovery rate from $18.8\%$ to $22.1\%$, eliminates the no-patch
failure mode entirely (baseline averages $\sim\!8$ no-patch failures
per run, ours averages $0$), and increases edit iteration on solved
trajectories from $4.4$ to $7.8$ \texttt{str\_replace} operations per
task. The \emph{multi-function} concentration of gain documented on
Verified is not visible on Lite because Lite contains no multi-file
tasks and only $54$ multi-function single-file tasks; the bulk of Lite
is single-function single-file tasks, on which the gain is
$+4.9$~pp ($19.5\%\!\to\!24.4\%$).

\begin{table}[h]
\centering
\caption{Trajectory-level behavioral metrics on SWE-Bench-Lite (14B,
R2E-Gym), run-means over three evaluation runs of each checkpoint.
Definitions match Table~\ref{tab:behavior_full}.}
\label{tab:behavior_lite}
\small
\setlength{\tabcolsep}{4pt}
\begin{tabular}{l c c c c c c c}
\toprule
& \multicolumn{2}{c}{Steps / task}
& \multicolumn{2}{c}{Edits / task}
& \multicolumn{2}{c}{Errors enc.}
& Recovery \\
\cmidrule(lr){2-3}\cmidrule(lr){4-5}\cmidrule(lr){6-7}
Setting & solved & unsolved & solved & unsolved & solved & unsolved & rate (\%) \\
\midrule
+ R2E-Gym
  & 17.5 & 21.3 & 4.4 & 5.2 & 4.8 & 5.4 & 18.8 \\
\textbf{+ FIM-Midtrain + R2E-Gym}
  & \textbf{24.6} & \textbf{31.7} & \textbf{7.8} & \textbf{11.1}
  & \textbf{5.7} & \textbf{7.5} & \textbf{22.1} \\
\midrule
\multicolumn{8}{l}{\textit{Output and context summaries}} \\
\midrule
& \multicolumn{2}{c}{Empty patch (\% failed)}
& \multicolumn{2}{c}{Loc.\ correct (\% all)}
& \multicolumn{2}{c}{Ctx @ success (tok)}
& Pass (\%) \\
\cmidrule(lr){2-3}\cmidrule(lr){4-5}\cmidrule(lr){6-7}
+ R2E-Gym
  & \multicolumn{2}{c}{$3.3$} & \multicolumn{2}{c}{$65.0$}
  & \multicolumn{2}{c}{$13{,}958$} & 18.0 \\
\textbf{+ FIM-Midtrain + R2E-Gym}
  & \multicolumn{2}{c}{$\mathbf{0.0}$} & \multicolumn{2}{c}{$\mathbf{71.0}$}
  & \multicolumn{2}{c}{$\mathbf{18{,}104}$} & \textbf{22.0} \\
\bottomrule
\end{tabular}
\end{table}

Failure-mode counts on Lite move in the same direction as on Verified:
the no-patch mode is eliminated ($\sim\!8\!\to\!0$ trajectories per
run), localization errors fall by about $11$, and patch errors rise by
about $7$, reflecting the same iterate-and-verify shift.

\begin{figure}[h]
\centering
\includegraphics[width=0.62\linewidth]{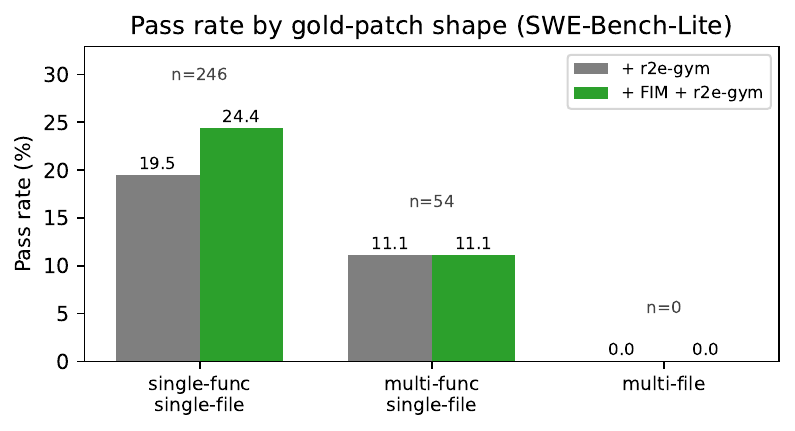}
\caption{Pass rate on SWE-Bench-Lite stratified by gold-patch shape,
averaged over three evaluation runs per checkpoint. Lite contains no
multi-file tasks; the multi-function single-file bucket is small
($n{=}54$) and shows no gain on this slice, with the $+4.0$~pp end-task
improvement coming entirely from the single-function single-file
bucket ($n{=}246$, $+4.9$~pp).}
\label{fig:failure_lite}
\end{figure}

\textbf{Per-instance head-to-head.}
Aggregating per-task outcomes across the three runs (run-majority vote
per checkpoint), the differential favors ours on Lite by a
$\sim\!1.8\times$ ratio overall and by a $\sim\!2.1\times$ ratio on the
single-function single-file bucket. On the multi-function single-file
bucket the differential is balanced, consistent with the small
absolute number of such tasks in Lite.


\end{document}